\documentclass{article}

\usepackage[preprint]{neurips_2026}
\makeatletter\renewcommand{\@noticestring}{}\makeatother

\usepackage[utf8]{inputenc} % allow utf-8 input
\usepackage[T1]{fontenc}    % use 8-bit T1 fonts
\usepackage{hyperref}       % hyperlinks
\usepackage{url}            % simple URL typesetting
\usepackage{booktabs}       % professional-quality tables
\usepackage{amsfonts}       % blackboard math symbols
\usepackage{amsmath}        % align, text in math, etc.
\usepackage{nicefrac}       % compact symbols for 1/2, etc.
\usepackage{microtype}      % microtypography
\usepackage{xcolor}         % colors
\usepackage{graphicx}       % for resizebox
\usepackage{multirow}       % for multirow in tables
\usepackage{makecell}       % for line breaks in table cells
\usepackage{colortbl}       % for row/cell colors in tables
\usepackage{wrapfig}        % for wraptable (text-wrapped tables)
\usepackage{todonotes}
\usepackage{listings}
\usepackage{pifont}
\lstdefinestyle{prompt}{%
  basicstyle=\ttfamily\footnotesize,
  breaklines=true,
  breakatwhitespace=true,
  columns=fullflexible,
  keepspaces=true,
  aboveskip=4pt,
  belowskip=4pt,
  frame=single,
  framesep=4pt,
  rulecolor=\color{black!30},
  xleftmargin=4pt,
  xrightmargin=4pt,
  showstringspaces=false,
}

\title{Bridging Modalities, Spanning Time: Structured Memory for Ultra-Long Agentic Video Reasoning}

\author{%
  \textbf{Jiazheng Li}$^{1}$, \textbf{Chi-Hao Wu}$^{2}$, \textbf{Yunze Liu}$^{2}$, \textbf{Kaize Ding}$^{3}$, \textbf{Jundong Li}$^{4}$, \textbf{Chuxu Zhang}$^{1}$ \\
  $^{1}$University of Connecticut \quad $^{2}$Memories.ai \quad $^{3}$Northwestern University \quad $^{4}$University of Virginia
}

\begin{document}

\maketitle

\renewcommand{\thefootnote}{}\footnotetext{Correspondence to \texttt{jiazheng.li@uconn.edu}, \texttt{chuxu.zhang@uconn.edu}.}\renewcommand{\thefootnote}{\arabic{footnote}}

\begin{abstract}
  Understanding ultra-long videos such as egocentric recordings, live streams, or
  surveillance footage spanning days to weeks, remains a challenge. For
  current multimodal LLMs: even with million-token context windows, frame
  budgets cover only tens of minutes of densely sampled video, and most
  evidence is discarded before inference begins. Memory-augmented and
  agentic approaches help with scale, but their retrieval remains
  fragmented across modalities and lacks long-range narrative summaries
  that span days or weeks. We propose \textbf{MAGIC-Video}, a training-free
  framework built around a \textbf{M}ultimod\textbf{A}l memory
  \textbf{G}raph with \textbf{I}nterleaved narrative \textbf{C}hain: the
  graph unifies episodic, semantic, and visual content through six typed
  edges and supports cross-modal retrieval, while the chain distils
  long-horizon entity biographies and recurring activity events.
  At inference time, an agentic loop
  interleaves graph retrieval with narrative fact injection, covering both
  the modality and time dimensions of ultra-long video in a single retrieval
  pipeline. On EgoLifeQA, Ego-R1 and MM-Lifelong, MAGIC-Video consistently
  outperforms strong general-purpose, long-video, and agentic baselines,
  with gains of 10.1, 7.4, and 5.9 points over the prior best agentic
  system on each benchmark.
  The code is publicly available: \href{https://github.com/lijiazheng0917/MAGIC-video}{\textcolor{magenta}{MAGIC-Video}}.
\end{abstract}

\section{Introduction}
\label{sec:intro}

Video understanding has progressed from short clips of seconds to
minutes~\citep{kinetics,msrvtt}, to medium-length content on the order of
tens of minutes~\citep{videomme,mlvu,longvideobench}, and more recently to
hour-scale videos~\citep{lvbench,hourvideo}. A further frontier is now
emerging: \emph{ultra-long} videos that span days, weeks, or even months.
EgoLifeQA~\citep{egolifeqa} ships \emph{7 days} of continuous
first-person footage, and MM-Lifelong~\citep{mmlifelong} extends to Day,
Week, and Month subsets, with its Month split reaching \emph{51 days} of
live-stream video. This regime is where many valuable real-world signals
reside, such as personal memory assistants over recorded life-logs,
agent activity logs accumulated over days, and long-term surveillance
or livestream recordings. The questions
they demand---how an entity evolves across days, which habits recur week
to week, or when a plan was made and when it was carried out---are
fundamentally temporal, entity-centric, and cross-modal.
Even frontier MLLMs with million-token
contexts~\citep{gemini31,gpt54} do not close the gap: a
million-token budget covers only tens of minutes of densely sampled
video, and at week scale any attempt to fit raw frames into the context
window forces aggressive downsampling, meaning that most evidence is discarded before
inference begins. An alternative is to build an external \emph{memory} of the video
offline and retrieve only the query-relevant evidence from it at
inference time, as in memory-augmented and agentic
pipelines~\citep{worldmm,m3agent,mmmem,egagent,avi,vgent,videorag,vimrag}.

Yet these memory-based systems, designed and evaluated on hour-scale
video, fall short of the cross-modal and long-horizon demands above
in two concrete ways.
\emph{First, cross-modal retrieval remains fragmented.}
Representative systems retrieve each modality through an independent
channel: WorldMM~\citep{worldmm} keeps three heterogeneous memory
stores (episodic, semantic, visual) and relies on an adaptive agent to
pick among them per query, while EGAgent~\citep{egagent} couples a
text-only entity graph with external visual-search tools. Even the
systems that place multimodal content in a single graph fall back on
similarity search over individual nodes, so a query such as ``who handed Jake the black
marker during the discussion?'', which needs to jointly activate a
visual clip and the semantic facts about the entities appearing in
it, cannot be resolved through any single retrieval step.
\emph{Second, bottom-up aggregation dilutes fine-grained details.} Representative systems build memory by summarising captions
level by level and extracting entities and triples per local window:
Vgent~\citep{vgent} retrieves over chunked multi-scale captions, and
VideoRAG~\citep{videorag} indexes caption summaries at successively
coarser granularities. As
granularity coarsens, individual mentions of an entity and individual
steps of an activity get abstracted away into generic window
descriptions, and the very details the query relies on are gone. No
\emph{top-down} pass scans the whole video to identify recurring
entities or multi-day activities and preserve their dated
moments as coherent retrieval units. A query such as ``how did the
cake-baking project unfold across the week?'' therefore either
retrieves coarse summaries that have lost the day-level detail, or
fine snippets that the reasoning backbone has to piece together on
its own.

\begin{figure}[t]
\centering
\includegraphics[width=\linewidth]{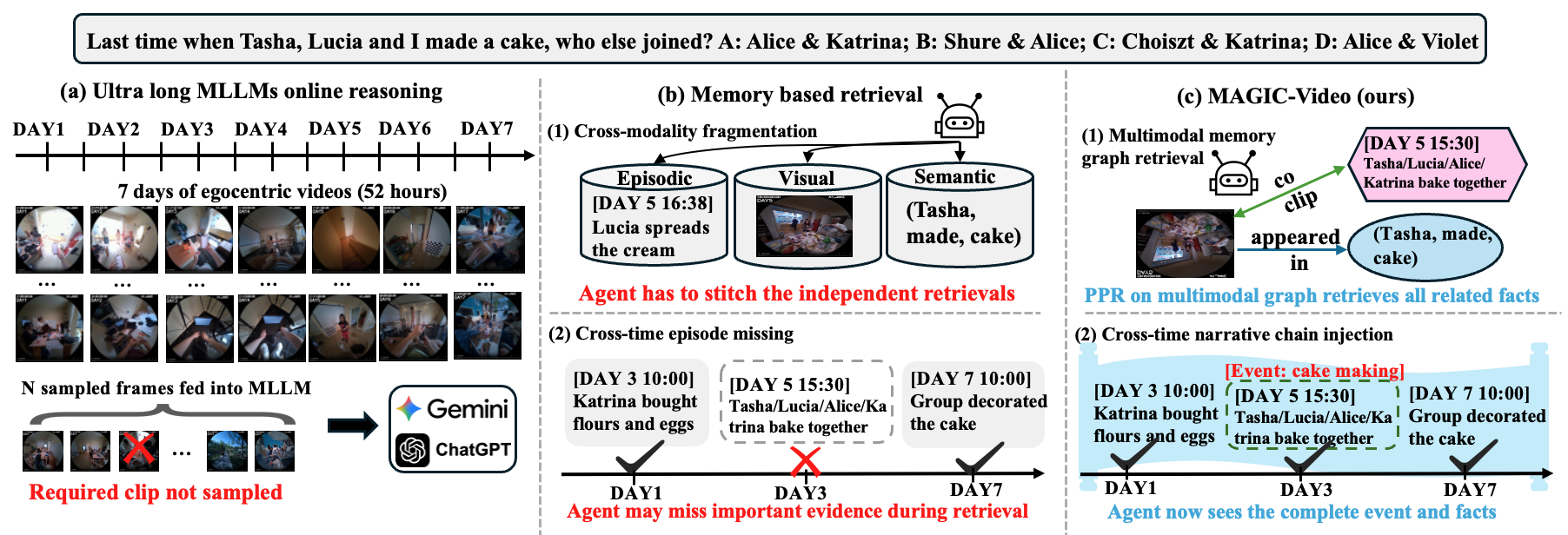}
\caption{Ultra-long video reasoning on EgoLifeQA \texttt{ID=244}.
\textbf{(a)} Long-context MLLMs sub-sample frames to fit a
million-token budget, dropping the required clip.
\textbf{(b)} Memory-based retrieval fails in two ways:
cross-modal fragmentation (each modality queried separately) and
missing cross-time episodes (bottom-up aggregations miss detailed facts). \textbf{(c)} MAGIC-Video fixes both via a Multimodal Memory
Graph (cross-modal PPR) and a Narrative Memory Chain (cross-time
fact injection).}
\label{fig:intro}
\vspace{-1cm}
\end{figure}

We therefore argue that the memory itself should be
\emph{multimodal at the structural level}, where visual, episodic,
and semantic evidence live in the same searchable space, and built
with
\emph{a top-down distillation} that complements bottom-up aggregation
by surfacing recurring entities and multi-day activities as
\emph{cross-time threads}---each thread a coherent unit the retriever
can pull in one shot. These two properties are complementary
(Figure~\ref{fig:intro}c): the cross-modal structure handles queries
about \emph{what is happening at a moment across modalities}, while
the cross-time threads handle queries about \emph{how situations
unfold over time}.

We realise these principles in MAGIC-Video, a training-free
\emph{structured-memory} framework that pairs two complementary
components. The
\emph{Multimodal Memory Graph} (Sec.~\ref{sec:mmg}) is a heterogeneous
graph over episode captions at multiple granularities, visual clips, named entities, and
consolidated semantic triples, connected by six typed cross-modal and
temporal edges; a single cross-modal Personalized PageRank pass
propagates query relevance across modality boundaries. The
\emph{Narrative Memory Chain} (Sec.~\ref{sec:nmc}) is the top-down
counterpart: an LLM applied offline scans the entire video to
distil per-entity biographies and multi-day activity events; two
families of cross-time threads that are then injected into retrieval
results through a hybrid lexical-semantic matching scheme. At inference time an
agentic loop (Sec.~\ref{sec:agentic_retrieval}) alternates between
search and answer, feeding the retrieved graph evidence together with
the injected narrative facts to the reasoning backbone.

On three ultra-long video benchmarks, MAGIC-Video consistently achieves
the best performance in its backbone class. On EgoLifeQA, MAGIC-Video surpasses the strongest prior system,
EGAgent~\citep{egagent} with Gemini 2.5 Pro, by 10.1 points.
On Ego-R1, it beats the previous best, WorldMM~\citep{worldmm}, by
7.4 points.
On MM-Lifelong, it outperforms the best prior agentic system,
ReMA-GPT-5~\citep{mmlifelong}, by 5.9 points.

\section{Related Work}
\label{sec:related}

\textbf{Long Video Understanding.} Frontier multimodal LLMs push
context windows to the million-token scale (e.g.,
Gemini~3.1~\citep{gemini31}, GPT-5.4~\citep{gpt54}), and video-specific
models such as LLaVA-Video~\citep{llavavideo},
VideoChat-Flash~\citep{videochatflash}, and Video-RTS~\citep{videorts}
further compress or reinforce temporal modelling. Even so, a
million-token budget covers only tens of minutes of densely sampled
video and forces aggressive downsampling on the days-to-weeks
horizons we target.

\textbf{Memory-Augmented and Agentic Video Reasoning.}
External-memory systems either keep unstructured or lightly
structured state (e.g., VideoMem~\citep{videomem}) or organise it
as graphs of entities (e.g., EGAgent~\citep{egagent},
WorldMM~\citep{worldmm}, Vgent~\citep{vgent}; see also
\citep{m3agent,mmmem,avi,videorag,vimrag,eventmemagent,longvideoagent}).
Some of these systems already place multimodal content on graph
nodes (e.g., M3-Agent stores text/visual/audio per entity), but
retrieval still proceeds through per-modality indices or per-tool
channels orchestrated by the agent, with no single graph-native
pass that propagates relevance across modalities. We instead
expose the multimodal evidence as first-class nodes connected by
typed cross-modal edges and retrieve over them in a single PPR pass. 

\textbf{Graph-Based Knowledge Retrieval.}
Graph-RAG systems retrieve via community summaries
(GraphRAG~\citep{graphrag}, LightRAG~\citep{lightrag}) or
Personalized PageRank (HippoRAG~\citep{hipporag}); follow-up work
adds temporal modelling (Zep~\citep{zep}, MemoTime~\citep{memotime})
or hierarchical agent memory (G-Memory~\citep{gmemory},
Optimus-1~\citep{optimus}, MAGMA~\citep{magma}). All operate over
purely textual nodes; we extend this line by admitting visual clips
as first-class graph nodes and running cross-modal PPR.

A more detailed comparison with each cited system is given in
Appendix~\ref{sec:related_extended}.

\section{Method}
\label{sec:method}

We now describe MAGIC-Video. The method has two phases: an
\textbf{offline} phase that processes each video once into a static
memory artefact, and an \textbf{online} phase that, given a question,
queries this artefact through a multi-round agentic loop.

\textbf{Offline}, we run a fixed preprocessing pipeline
(Sec.~\ref{sec:preprocessing}) that turns the raw video into
multi-granularity captions, named-entity annotations, semantic
triples, and visual embeddings. From these artefacts we build the
two complementary components of our \emph{structured memory}: a
\textbf{Multimodal Memory Graph}
(MMG, Sec.~\ref{sec:mmg}) that places episodic captions, visual
clips, named entities, and semantic triples into a single
heterogeneous graph connected by six typed edges, and a
\textbf{Narrative Memory Chain} (NMC, Sec.~\ref{sec:nmc}) that
distils long-horizon topic chains and cross-time event chains from
the captions via an LLM.

\textbf{Online}, given a question $q$ and an optional query time
$t$, an \textbf{agentic retrieval} procedure
(Sec.~\ref{sec:agentic_retrieval}) runs cross-modal Personalized
PageRank over the time-filtered subgraph $\mathcal{G}_{\le t}$,
attaches matching narrative facts from NMC, and feeds the combined
context to a reasoning backbone that decides whether to issue another
search or commit to an answer.

\begin{figure}[t]
\centering
\includegraphics[width=\linewidth]{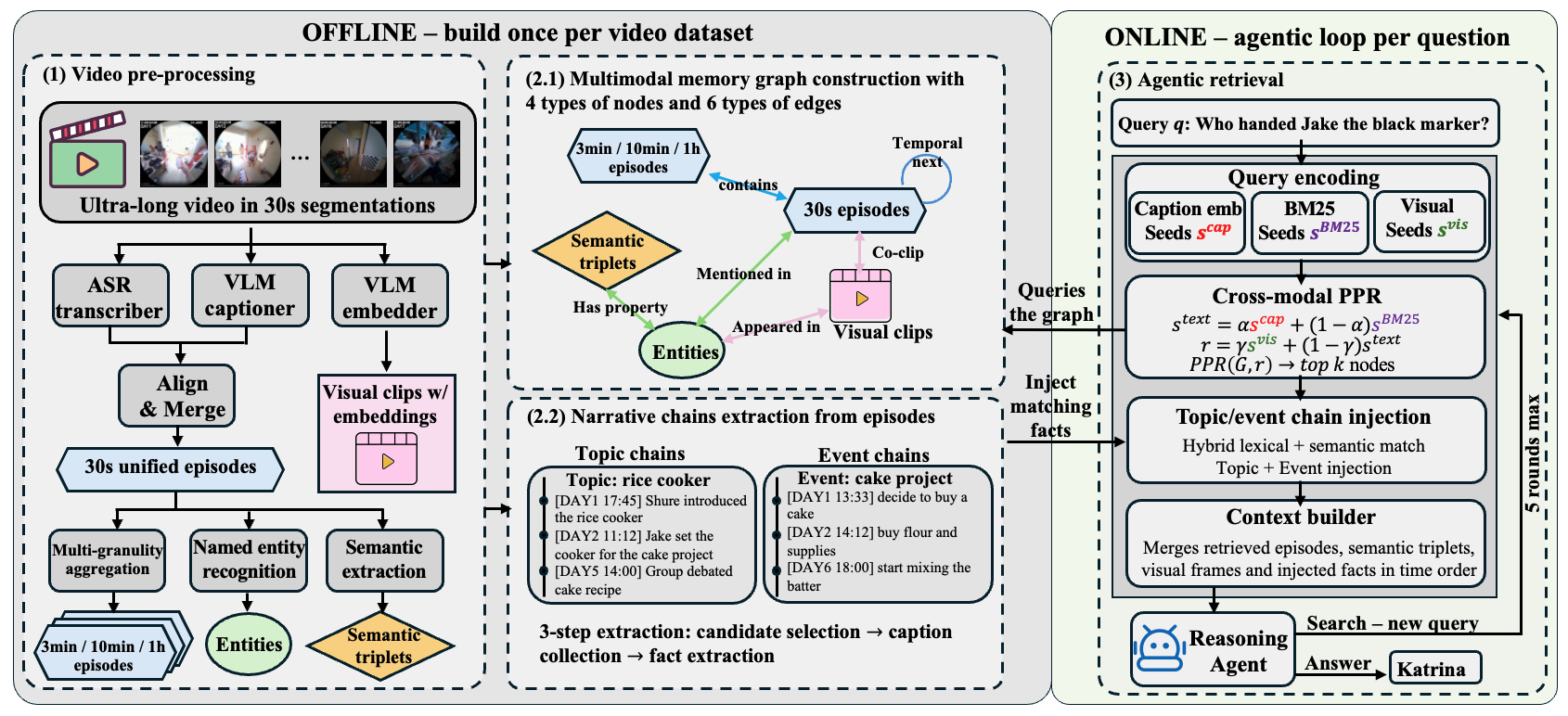}
\caption{MAGIC-Video pipeline.
\textbf{Offline (left):} preprocessing produces multi-granularity
captions, named entities, semantic triples, and visual embeddings,
from which we build the \textbf{Multimodal Memory Graph} (four
node types connected by six typed edges) and the
\textbf{Narrative Memory Chain} (topic chains + event chains).
\textbf{Online (right):} for each question, an agentic loop seeds
cross-modal Personalized PageRank over the graph, injects matching chains, and feeds the merged context to the reasoning backbone, which
either refines its search or commits to an answer.}
\label{fig:overview}
\end{figure}

\subsection{Preprocessing}
\label{sec:preprocessing}

Preprocessing (Fig.~\ref{fig:overview}, block~(1)) turns a raw long
video into the five artefacts the rest of the pipeline consumes:
multi-granularity episodic captions at 30\,s / 3\,min / 10\,min /
1\,h, named-entity annotations, consolidated semantic triples,
per-clip visual embeddings, and a BM25 index over the
captions. All five are produced by chaining ASR and a visual-caption
model with an LLM-based merge, aggregation, and
OpenIE~\citep{openie} pipeline. Specific model
choices and prompt templates are listed in
Appendix~\ref{sec:impl_details} and~\ref{sec:prompts}; per-stage
operation counts are reported in Sec.~\ref{sec:efficiency}.

\subsection{Multimodal Memory Graph Construction}
\label{sec:mmg}

From the preprocessing artefacts of Sec.~\ref{sec:preprocessing} we
now build the Multimodal Memory Graph, a heterogeneous graph
$\mathcal{G} = (\mathcal{V}, \mathcal{E})$ that places all per-video
evidence into a single retrieval substrate.

\textbf{Schema.} The graph has four node types ---
\emph{Episode} (multi-granularity captions: 30\,s / 3\,min / 10\,min / 1\,h),
\emph{Visual-clip} (per-clip visual embedding),
\emph{Entity} (NER mention; first-person canonicalised), and
\emph{Semantic} (consolidated triple) --- and six typed edges:
\textsc{mentioned\_in} (entity\,$\to$\,episode that names it);
\textsc{co\_clip} (episode\,---\,same-window visual clip);
\textsc{appeared\_in} (entity\,$\to$\,visual clip; a shortcut for
\textsc{mentioned\_in}\,$\to$\,\textsc{co\_clip});
\textsc{has\_property} (entity\,$\to$\,triple it appears in as
subject or object);
\textsc{temporal\_next} (adjacent same-granularity episodes,
with time-decayed weight); and
\textsc{contains} (coarse episode\,$\to$\,finer episode it covers).
This vocabulary gives MMG cross-modal shortcuts in a single
traversal: textual seeds reach visual clips via
\textsc{mentioned\_in}\,$\to$\,\textsc{co\_clip}, and visual seeds
reach semantic facts via
\textsc{appeared\_in}\,$\to$\,\textsc{has\_property}.

\textbf{Construction.} The graph is instantiated in a single
offline pass (per-benchmark node/edge counts in
Appendix~\ref{sec:graph_stats}). Text embeddings are computed for
episode and semantic nodes only; visual-clip and entity nodes have
their embeddings zeroed and serve as PageRank \emph{bridges}, with
visual content arriving through the separate
$\mathbf{s}^{\mathrm{vis}}$ seed channel of
Sec.~\ref{sec:agentic_retrieval}. Edges carry typed weights that
down-weight derived shortcuts and apply a $1/(\Delta t + 1)$ decay
on \textsc{temporal\_next} (full schedule in
Appendix~\ref{sec:edge_weights}). The schema is visualised in the
offline panel of Figure~\ref{fig:overview}.

\subsection{Narrative Memory Chain Extraction}
\label{sec:nmc}

While MMG supports local, caption-level retrieval, many life-long
questions ask about what happens \emph{between} retrieved
captions---how an entity's state changes, or how an activity recurs
across weeks. NMC distils the captions offline into two
complementary structures: \emph{topic chains} that trace individual
entities, and \emph{event chains} that link related activity steps.
Both are produced by feeding captions of progressively finer
granularity to an LLM.

\textbf{Topic chains} trace the lifecycle of a single entity. We
select entities that recur as subject or object in the semantic
triples, gather every 30-second
caption that mentions them, and prompt an LLM for meaningful,
person-grounded, action-bearing facts about each entity. The extraction prompt is given in
Appendix~\ref{sec:prompt_topic_chain}. The output for each topic
$e$ is a chronologically ordered list
$\mathcal{N}_e = \{(t_i, f_i)\}_i$ of timestamped facts.
For example, the second fact of the topic chain $e
= \texttt{hot pot}$ is rendered to the controller as
\emph{``[Topic: hot pot] [DAY1 17:42] Shure handed hot pot meat to
JAKE in the store.''}

\textbf{Event chains} link related activity steps across time. We
extract per-day activities from 10-minute captions, identify
activities that are related across time (recurring routines,
progressive project stages, or intermittent work) into dated chains,
and enrich each step from the 30-second captions inside its time
range. The corresponding extraction prompts are given in
Appendices~\ref{sec:prompt_event_chain_12}
and~\ref{sec:prompt_event_chain_3}. Each event chain $k$ is a list
of dated steps $\mathcal{S}_k = \{(t_j^{\mathrm{start}},
t_j^{\mathrm{end}}, d_j)\}_j$ together with a set of key entities.
For example, the second step
of the chain \texttt{Cake and dessert preparation} is
rendered to the controller as \emph{``[Event: Cake and dessert
preparation] [DAY3 22:00:00 -- 22:48:00] Lucia guided JAKE in
spreading frosting while Tasha smoothed the sides and placed
strawberries.''}

\subsection{Agentic Retrieval}
\label{sec:agentic_retrieval}

At inference time, MAGIC-Video exposes MMG and NMC to the reasoning
backbone through a multi-round retrieval loop that alternates
between search and answer.

\textbf{Cross-modal PPR over MMG.}
Given a search query $q$, MAGIC-Video runs one cross-modal
PPR pass over MMG. Following HippoRAG's~\citep{hipporag}
seed-weighted PPR template, we build the PPR reset vector
$\mathbf{r} \in \mathbb{R}^{|\mathcal{V}|}$ from three complementary
seed channels driven by $q$:
\begin{equation}
  \mathbf{s}^{\mathrm{text}}(q) \;=\; \alpha\,\mathbf{s}^{\mathrm{cap}}(q)
  + (1-\alpha)\,\mathbf{s}^{\mathrm{BM25}}(q),
  \qquad
  \mathbf{r} \;=\; \gamma\,\mathbf{s}^{\mathrm{vis}}(q)
  \;+\; (1-\gamma)\,\mathbf{s}^{\mathrm{text}}(q),
  \label{eq:reset}
\end{equation}
where $\alpha$ trades caption-embedding against BM25 within the
textual modality, and $\gamma$ trades the textual seed against
$\mathbf{s}^{\mathrm{vis}}$, restricted to visual-clip nodes
($q$ is re-encoded online by VLM2Vec's text-side into the same
3584-d space, keeping both channels intra-encoder). All three
seed scores are max-normalized to $[0,1]$ before mixing and
$\mathbf{r}$ is finally normalized to sum to $1$.
Setting $\gamma = 0$ disables the visual channel and reduces
$\mathbf{r}$ to the text-only mix $\mathbf{s}^{\mathrm{text}}$,
matching HippoRAG's seed-vector pattern over textual nodes;
MAGIC-Video extends it with the cross-modal visual term. We run undirected Personalized
PageRank~\citep{page1999pagerank,jeh2003scaling}
$\boldsymbol{\pi} = d\,\tilde{A}^\top\boldsymbol{\pi} + (1-d)\,\mathbf{r}$
on the edge-weighted subgraph and rerank by
\begin{equation}
  \mathrm{score}(v) \;=\; \pi(v) \cdot \tfrac{\cos(\mathbf{q}, \mathbf{e}_v) + 1}{2},
\end{equation}
which breaks ties between structurally close but semantically
irrelevant nodes; visual-clip and entity nodes carry no text
embedding (Sec.~\ref{sec:mmg}), so we set $\cos(\mathbf{q},\mathbf{0}){=}0$
by convention and they rank purely by $\pi(v)$. A multi-scale filter then caps retained nodes per
granularity and node type (quotas in
Appendix~\ref{sec:hyperparams}).

\textbf{Narrative injection from NMC.}
Alongside the PPR pass, NMC's topic and event chains are matched
against the same search query $q$ in two tiers and appended to
the retrieved set: \emph{Tier 1} admits chains whose keyword
entities appear in $q$; \emph{Tier 2} admits chains by cosine similarity between $q$ and the chain's content.
Admitted facts are time-filtered by $t$ and deduplicated against
retrieved episodes. Full procedure (candidate anchoring, per-tier
budgets, thresholds) is in Appendix~\ref{sec:nmc_injection}.

\textbf{Context formatting and answering.}
The merged context fed to the reasoning backbone is the time-ordered
union of all retrieved items across rounds:
\begin{equation}
  \mathcal{C} \;=\; \text{TimeOrdered}\!\left(
  \textstyle\bigcup_{r=1}^{R}\,
  \mathcal{P}_r \cup \mathcal{F}_r \cup \mathcal{T}_r \cup \mathcal{H}_r
  \right),
  \label{eq:context}
\end{equation}
where $\mathcal{P}_r$, $\mathcal{F}_r$, $\mathcal{T}_r$,
$\mathcal{H}_r$ are the episode captions (passages), visual frames,
semantic triples, and NMC narrative facts retrieved in round $r$. Any of
these sets may be empty in a given round (e.g., no visual clip is
retrieved, or no chain matches the query). Each retained item is
rendered with an explicit type tag (template in
Appendix~\ref{sec:prompts}).

\textbf{Multi-round loop.}
The two retrieval primitives above are composed by an agentic
controller. In each round, the controller is given the question and
the evidence retrieved so far, and emits either \texttt{search}
(a natural-language query $q$) or \texttt{answer} (stop and
commit). A \texttt{search} triggers one PPR pass over MMG plus
one matching step on NMC; the new evidence is deduplicated against
previous rounds by node id and appended to the running context. The loop terminates
when the controller emits \texttt{answer} or after $R = 5$ search
rounds. On termination, the reasoning backbone reads the merged
context $\mathcal{C}$ end-to-end and produces the final answer
directly.

\begin{table}[t]
\caption{Results on the EgoLifeQA and Ego-R1 benchmarks.
Entries marked with $*$ are taken from the original papers; all other numbers
are reproduced by us under identical preprocessing.
\textbf{Bold} marks the best result and \underline{underline} marks the
second-best.}
\label{tab:main_results}
\centering
\resizebox{\textwidth}{!}{%
\begin{tabular}{l cc cccccc ccc}
\toprule
\multirow{2}{*}{Model} & \multirow{2}{*}{\# Frames} & \multirow{2}{*}{Modality} & \multicolumn{6}{c}{EgoLifeQA} & \multicolumn{3}{c}{Ego-R1} \\
\cmidrule(lr){4-9} \cmidrule(lr){10-12}
& & & EL & ER & HI & RM & TM & Avg. & Manual & Gemini & Avg. \\
\midrule
\rowcolor{gray!15} \multicolumn{12}{c}{\textbf{General MLLMs}} \\
 & 64 & V & 32.8 & 30.2 & 45.9 & 28.8 & 20.6 & 31.2 & 24.0 & 42.7 & 33.3 \\
Qwen3.5-9B & 512 & T & 25.6 & 28.6 & 47.5 & 35.2 & 41.3 & 33.4 & 33.3 & 46.7 & 40.0 \\
 & 64 & V+T & 28.8 & 31.7 & 45.9 & 33.6 & 36.5 & 33.8 & 22.7 & 64.0 & 43.3 \\
\midrule
Qwen3.5-Flash & 64 & V+T & 42.4 & 39.7 & 45.9 & 39.2 & 46.0 & 41.8 & 30.7 & 60.0 & 45.3 \\
GPT-5.4 Mini & 64 & V+T & 37.6 & 40.5 & 44.3 & 34.4 & 41.3 & 38.8 & 22.7 & 40.0 & 31.3 \\
Gemini 3.1 Flash Lite & 64 & V+T & 44.0 & 41.3 & 42.6 & 47.2 & 38.1 & 43.2 & 29.3 & \underline{66.7} & 48.0 \\
GPT-4.1* & -- & T & 32.0 & 39.7 & 39.3 & 32.8 & 39.7 & 36.0 & -- & -- & -- \\
Gemini 2.5 Pro* & 3000 & V+T & 45.6 & 48.4 & 51.7 & 41.6 & 52.4 & 46.8 & -- & -- & -- \\
\midrule
\rowcolor{gray!15} \multicolumn{12}{c}{\textbf{Long Video MLLMs}} \\
VideoLLaMA3-7B & 128 & V & 32.8 & 35.7 & 37.7 & 27.2 & 33.3 & 32.8 & 34.7 & 36.0 & 35.3 \\
InternVideo2.5-8B & 512 & V & 34.4 & 37.3 & 42.6 & 30.4 & 31.7 & 34.8 & 28.0 & 21.3 & 24.7 \\
LongVA-7B & 128 & V & 33.6 & 38.1 & 36.1 & 39.2 & 28.6 & 35.8 & 28.0 & 18.7 & 23.3 \\
VideoChat-Flash-7B & 1024 & V & 36.8 & 39.7 & 34.4 & 31.2 & 41.3 & 36.4 & 33.3 & 34.7 & 34.0 \\
\midrule
\rowcolor{gray!15} \multicolumn{12}{c}{\textbf{RAG-based Video MLLMs}} \\
LLaVA-Video-7B + Video-RAG* & 64 & V & -- & -- & -- & -- & -- & 30.0 & -- & -- & 29.3 \\
LongVA-7B + Video-RAG* & 64 & V & -- & -- & -- & -- & -- & 26.0 & -- & -- & 31.0 \\
\midrule
\rowcolor{gray!15} \multicolumn{12}{c}{\textbf{Agentic Video LLMs}} \\
EgoButler-GPT-4o* & -- & T & 34.4 & 42.1 & 29.5 & 30.4 & 44.4 & 36.2 & -- & -- & -- \\
EgoButler-Gemini 1.5 Pro* & -- & T & 36.0 & 37.3 & 45.9 & 30.4 & 34.9 & 36.9 & -- & -- & -- \\
EGAgent-Gemini 2.5 Pro* & 1FPS$\rightarrow$50 & V+T & \underline{54.4} & \underline{57.1} & \underline{60.3} & \underline{62.4} & \textbf{74.6} & \underline{57.5} & -- & -- & -- \\
VideoAgent* & 1FPS$\rightarrow$8 & V & -- & -- & -- & -- & -- & 29.2 & -- & -- & 39.6 \\
LLaVA-OneVision + T* & 1FPS$\rightarrow$8 & V+T & -- & -- & -- & -- & -- & 35.4 & -- & -- & 35.6 \\
EgoR1-Qwen2.5-3B* & 1FPS & V+T & -- & -- & -- & -- & -- & 36.0 & -- & -- & 46.0 \\
WorldMM-Qwen3.5-Flash & Full & V+T & 49.6 & 54.8 & 54.1 & \underline{62.4} & \underline{60.3} & 56.0 & \underline{48.0} & \underline{66.7} & \underline{57.3} \\
\rowcolor{blue!10} MAGIC-Video-Qwen3.5-Flash & Full & V+T & \textbf{67.2} & \textbf{65.9} & \textbf{67.2} & \textbf{66.4} & \textbf{74.6} & \textbf{67.6} & \textbf{50.7} & \textbf{78.7} & \textbf{64.7} \\
\bottomrule
\end{tabular}%
}
\end{table}

\section{Experiments}
\label{sec:experiments}

\subsection{Experimental Setup}
\label{sec:setup}

\textbf{Benchmarks.}
We evaluate on three question sets drawn from two ultra-long video
corpora. \textbf{EgoLifeQA}~\citep{egolifeqa} and \textbf{Ego-R1}
Bench~\citep{egor1} use the week-long egocentric
life-logging recordings released by~\citet{egolifeqa}; following prior
work~\citep{egagent,worldmm}, we evaluate on the A1\_JAKE subject,
whose recording consists of 51.9\,hours of continuous first-person
footage over 7 days. The two benchmarks share this video but differ
in their question sets: EgoLifeQA provides 500 multiple-choice
questions grouped into five subtasks---EntityLog (EL), EventRecall
(ER), HabitInsight (HI), RelationMap (RM), and TaskMaster (TM)---each
probing a distinct kind of long-range reasoning, while Ego-R1 Bench
provides 50 questions split between a human-authored \emph{Manual} set and a
model-generated \emph{Gemini} set.
\textbf{MM-Lifelong}~\citep{mmlifelong} covers three temporal
scales (Day / Week / Month); we focus on the \emph{Month} subset, the
longest-horizon scale, consisting of 105.6\,hours of live-stream video
spanning 51 days. It contains 623 open-ended questions across
eleven categories---Counting, Entity Recognition, Causal
Reasoning, Temporal Reasoning, Event Recognition, Language Content
Recall, Hallucination Detection, Attribute Recognition, Social
Interaction, State Change, Event Tracking.

\textbf{Metrics.}
For EgoLifeQA and Ego-R1 we report multiple-choice accuracy. MM-Lifelong is
scored with an LLM-as-judge protocol; we follow the original
paper~\citep{mmlifelong} and refer readers there for full scoring details.
Unless otherwise noted, all MM-Lifelong numbers in
Table~\ref{tab:mm_lifelong} use \textit{GPT-5 as the judge} to ensure a
consistent comparison across all methods; to additionally assess robustness
to the choice of judge, we report results under two other judges in
Table~\ref{tab:mml_judge_comparison}.

\textbf{Baselines.}
We compare against four categories of baselines.\footnote{On MM-Lifelong we
deliberately avoid the GPT-5.4 and Gemini 3.1 versions used on
EgoLifeQA/Ego-R1 and instead use earlier releases (GPT-5 Mini, Gemini 3
Flash), because the live-stream videos in MM-Lifelong may overlap with the
training data of the newer models; EgoLifeQA and Ego-R1, built from a
controlled in-house recording, are not affected by this concern.}
(1) \emph{General MLLMs} include Qwen3.5-9B~\citep{qwen35}, Qwen3.5-Flash,
GPT-5.4 Mini~\citep{gpt54}, GPT-5 Mini~\citep{gpt54}, GPT-4.1~\citep{gpt41},
Gemini 3 Flash~\citep{gemini31}, Gemini 3.1 Flash Lite~\citep{gemini31},
and Gemini 2.5 Pro~\citep{gemini25}, evaluated with uniform frame
sampling.
(2) \emph{Long Video MLLMs} include VideoLLaMA3-7B~\citep{videollama3},
InternVideo2.5-8B~\citep{internvideo25}, LongVA-7B~\citep{longva}, and
VideoChat-Flash-7B~\citep{videochatflash}.
(3) \emph{RAG-based Video MLLMs} wrap LLaVA-Video-7B~\citep{llavavideo}
and LongVA-7B~\citep{longva} with Video-RAG~\citep{videoraglocal}.
(4) \emph{Agentic Video MLLMs} include EgoButler~\citep{egolifeqa},
EGAgent~\citep{egagent}, VideoAgent~\citep{videoagent},
EgoR1~\citep{egor1}, and WorldMM~\citep{worldmm} for EgoLifeQA/Ego-R1,
and VideoMind~\citep{videomind}, LongVT~\citep{longvt},
DeepVideoDiscovery~\citep{dvd}, and ReMA~\citep{mmlifelong} for
MM-Lifelong.

\textbf{Models and hyperparameters.}
MAGIC-Video uses \texttt{gpt-oss-120b}~\citep{gptoss} for offline
preprocessing and as the online retrieval controller, and
Qwen3.5-Flash~\citep{qwen35} as the reasoning backbone. For fairness,
we use the same hyperparameters across all experiments wherever
applicable. At query time $t$, we extract a time-filtered subgraph
$\mathcal{G}_{\le t} \subseteq \mathcal{G}$ retaining only nodes
visible at or before $t$. Full model list, captioning
pipeline, and complete hyperparameter values are given in
Appendices~\ref{sec:impl_details} and~\ref{sec:hyperparams}.

\begin{table}[t]
\caption{Results on the MM-Lifelong benchmark. Entries marked with $*$ are
taken from the original papers; all other numbers are reproduced by us under
identical preprocessing.
\textbf{Bold} marks the best result and \underline{underline} marks the
second-best.}
\label{tab:mm_lifelong}
\centering
\resizebox{\textwidth}{!}{%
\begin{tabular}{l cc ccccccccccc c}
\toprule
Model & \# Frames & Modality & \makecell{Cnt.\\{\scriptsize 135}} & \makecell{ER\\{\scriptsize 108}} & \makecell{CR\\{\scriptsize 107}} & \makecell{TR\\{\scriptsize 72}} & \makecell{EvR.\\{\scriptsize 54}} & \makecell{LCR\\{\scriptsize 46}} & \makecell{HD\\{\scriptsize 44}} & \makecell{AR\\{\scriptsize 26}} & \makecell{SI\\{\scriptsize 23}} & \makecell{SC\\{\scriptsize 5}} & \makecell{ET\\{\scriptsize 3}} & Avg. \\
\midrule
\rowcolor{gray!15} \multicolumn{15}{c}{\textbf{General MLLMs}} \\
 & 64 & V & 8.1 & 5.6 & 7.0 & 12.5 & 9.3 & 13.0 & 13.6 & 7.7 & 2.2 & \textbf{10.0} & 0.0 & 8.6 \\
Qwen3.5-9B & 512 & T & 5.2 & \underline{25.9} & 14.0 & 12.5 & 25.0 & \underline{25.0} & 21.6 & 25.0 & \textbf{17.4} & 0.0 & 0.0 & 16.7 \\
 & 64 & V+T & 7.8 & 24.5 & 14.0 & 11.1 & \underline{27.8} & \textbf{33.7} & \underline{23.9} & \textbf{32.7} & \underline{13.0} & 0.0 & 0.0 & 18.1 \\
\midrule
Qwen3.5-Flash & 64 & V+T & 4.1 & 17.1 & 12.6 & 10.4 & 8.3 & 9.8 & 21.6 & 15.4 & 8.7 & 0.0 & 0.0 & 11.2 \\
GPT-5 Mini & 64 & V+T & \underline{11.9} & 15.3 & 15.4 & 6.2 & 11.1 & 9.8 & \textbf{25.0} & \underline{26.9} & \underline{13.0} & 0.0 & 0.0 & 13.6 \\
Gemini 3 Flash & 64 & V+T & 4.1 & 19.9 & \underline{21.0} & \underline{13.9} & 21.3 & 17.4 & 9.1 & 19.2 & \underline{13.0} & 0.0 & 0.0 & 14.6 \\
\midrule
\rowcolor{gray!15} \multicolumn{15}{c}{\textbf{Long Video MLLMs}} \\
VideoLLaMA3-7B & 128 & V & 6.3 & 4.6 & 4.2 & 11.8 & 6.5 & 5.4 & 3.4 & 15.4 & 4.3 & 0.0 & 0.0 & 6.3 \\
InternVideo2.5-8B & 512 & V & \textbf{12.6} & 7.4 & 7.0 & 6.9 & 15.7 & 8.7 & 9.1 & 7.7 & 4.3 & 0.0 & 0.0 & 9.1 \\
LongVA-7B & 128 & V & 3.0 & 4.2 & 6.1 & \underline{13.9} & 11.1 & 9.8 & 6.8 & 7.7 & 6.5 & 0.0 & 0.0 & 6.7 \\
VideoChat-Flash-7B & 1024 & V & 8.5 & 8.3 & 7.9 & \underline{13.9} & 13.0 & 9.8 & 6.8 & 7.7 & 4.3 & 0.0 & 0.0 & 9.1 \\
\midrule
\rowcolor{gray!15} \multicolumn{15}{c}{\textbf{Agentic Video MLLMs}} \\
VideoMind-7B* & Full & V+T & -- & -- & -- & -- & -- & -- & -- & -- & -- & -- & -- & 8.4 \\
LongVT-7B* & Full & V+T & -- & -- & -- & -- & -- & -- & -- & -- & -- & -- & -- & 7.5 \\
DeepVideoDiscovery-7B* & Full & V+T & -- & -- & -- & -- & -- & -- & -- & -- & -- & -- & -- & 10.6 \\
ReMA-GPT-5* & Full & V+T & -- & -- & -- & -- & -- & -- & -- & -- & -- & -- & -- & \underline{18.6} \\
\rowcolor{blue!10} MAGIC-Video-Qwen3.5-Flash & Full & V+T & 7.4 & \textbf{31.0} & \textbf{29.4} & \textbf{27.1} & \textbf{36.1} & \textbf{40.2} & 15.9 & \textbf{32.7} & \textbf{17.4} & \textbf{10.0} & 0.0 & \textbf{24.5} \\
\bottomrule
\end{tabular}%
}
\vspace{-0.5cm}
\end{table}

\subsection{Main Results}
\label{sec:main_results}

\textbf{Consistent improvements across benchmarks.}
MAGIC-Video achieves the highest overall accuracy on every
benchmark, despite running on a Qwen3.5-Flash reasoning backbone.
On EgoLifeQA (Table~\ref{tab:main_results}) it reaches 67.6\%
average, surpassing the strongest prior system, EGAgent (Gemini
2.5 Pro backbone), by 10.1 points; even with the much smaller
backbone, the gap over the best single-shot frontier MLLM,
Gemini~2.5~Pro at 3000 frames ($46.8$\%), widens to $+20.8$. On
Ego-R1 it reaches 64.7\% average, surpassing the previous best
system, WorldMM (matched Qwen3.5-Flash backbone), by 7.4 points.
On MM-Lifelong (Table~\ref{tab:mm_lifelong}) it reaches 24.5\%
under the GPT-5 judge, surpassing the best prior agentic system,
ReMA-GPT-5, by 5.9 points, while the four 7--8B Long-Video MLLMs
cluster at 6.3--9.1\%, leaving an almost $3\times$ gap at
comparable backbone scale. The ranking is preserved when scoring
MM-Lifelong with two alternative judges
(Table~\ref{tab:mml_judge_comparison} in the appendix): MAGIC-Video
scores 28.7 / 29.4 / 24.5 with Qwen3.5-Flash / GPT-5 Mini / GPT-5,
versus at most 20.0 / 20.9 / 18.1 for the strongest General MLLM
baseline, indicating that the observed advantage is not a
judge-specific artefact.

\textbf{Cross-modal and long-horizon subtasks benefit the most.}
On EgoLifeQA, the largest gains over WorldMM appear on EntityLog
($+17.6$), TaskMaster ($+14.3$), HabitInsight ($+13.1$), and
EventRecall ($+11.1$)---the subtasks whose answers most directly
require linking a text mention to a specific visual moment or spanning
multiple days, which Sec.~\ref{sec:mmg_analysis} traces back to MMG's
typed cross-modal edges together with NMC's cross-day narrative facts.
RelationMap shows the smallest matched-backbone gap ($+4.0$),
consistent with these queries being largely textually recoverable
from the IR baseline's top-$k$ episodes already.
On MM-Lifelong the pattern is even sharper: MAGIC-Video more than
doubles the best Long Video MLLM on Causal Reasoning ($29.4$ vs.\
$7.9$) and Event Recognition ($36.1$ vs.\ $15.7$), and improves
Temporal Reasoning from $13.9$ to $27.1$. These are precisely the
categories that require reasoning across multiple days rather than
within a single retrieved snippet, where NMC's pre-distilled entity
biographies and event chains supply the missing cross-day
structure (Sec.~\ref{sec:nmc_analysis}). Counting (7.4 vs.\
InternVideo2.5's 12.6) and Hallucination Detection (15.9 vs.\ GPT-5
Mini's 25.0) are the two categories where MAGIC-Video does not
lead, because both reward dense single-pass frame coverage over
multi-round retrieval --- a trade-off we revisit in
Sec.~\ref{sec:limitations}.

\section{In-Depth Analysis}
\label{sec:analysis}

\subsection{Ablation Study: Cumulative Effect of MMG and NMC}
\label{sec:ablation}

\begin{figure}[ht]
\centering
\includegraphics[width=0.55\textwidth]{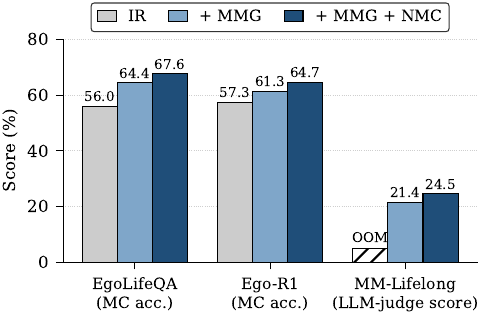}
\caption{Cumulative ablation of stacking MMG and
then NMC on top of an independent-retrieval (IR) baseline. The
third bar is the full MAGIC-Video system.
EgoLifeQA / Ego-R1 use MC accuracy; MM-Lifelong uses GPT-5-judged
answer accuracy (same 0--100 scale).}
\label{fig:ablation}
\end{figure}

\textbf{Setup.} Figure~\ref{fig:ablation} adds the two
structured-memory components (MMG and NMC) on top of a WorldMM-style
independent-retrieval baseline: row 2
replaces the three per-modality indices with the Multimodal Memory
Graph, and row 3 additionally enables Narrative Memory Chain injection.

\textbf{Effect.} The Multimodal Memory Graph adds $+8.4$ on
EgoLifeQA, $+4.0$ on Ego-R1, and on MM-Lifelong restores a working
retriever at $21.4$ where the independent baseline OOMs (per-granularity HippoRAG indices exhaust host RAM; Appendix~\ref{sec:ir_oom}). The Narrative Memory Chain adds another
$+3.2$ / $+3.4$ / $+3.1$ to reach $67.6$ / $64.7$ / $24.5$. Neither
component alone reaches the full system, so the two are complementary
rather than redundant.

\subsection{Impact of the Multimodal Memory Graph}
\label{sec:mmg_analysis}

We now drill into \emph{where} MMG's $+8.4$-point EgoLifeQA lift
concentrates, by breaking it down across all five subtask categories
that make up EgoLifeQA's evaluation taxonomy
(Table~\ref{tab:mmg_subtask}).
\begin{table}[ht]
\centering
\small
\caption{Per-subtask effect of the Multimodal Memory Graph on
EgoLifeQA. \emph{IR} is the independent-retrieval baseline used in
Figure~\ref{fig:ablation}; \emph{+ MMG} is MAGIC-Video with the multimodal memory graph
but without chain injection.}
\label{tab:mmg_subtask}
\begin{tabular}{lrrrr}
\toprule
Subtask & $n$ & IR & + MMG & $\Delta$ \\
\midrule
EL & 125 & 49.6 & 61.6 & +12.0 \\
ER & 126 & 54.8 & 65.9 & +11.1 \\
HI &  61 & 54.1 & 60.7 & +6.6 \\
RM & 125 & 62.4 & 65.6 & +3.2 \\
TM &  63 & 60.3 & 68.3 & +7.9 \\
\midrule
\textbf{Avg.} & 500 & 56.0 & 64.4 & +8.4 \\
\bottomrule
\end{tabular}
\end{table}

\textbf{Mechanism.} MMG's largest gains concentrate on subtasks
where text-to-visual grounding is required: EntityLog
($+12.0$\,pp) and EventRecall ($+11.1$\,pp), both of which need the
typed-edge path \textsc{mentioned\_in}\,$\to$\,\textsc{co\_clip} to
co-retrieve a text mention with its visual moment in one PPR pass.
The smaller RelationMap lift ($+3.2$\,pp) reflects a ceiling
effect: relational queries between persons are often recoverable
from text alone, so the IR baseline is already strong ($62.4\%$).
Retrieved contexts are a non-trivial mix of Episode ($63$--$74\%$),
Semantic-triple ($21$--$35\%$), and Visual-frame ($2$--$10\%$) items
across the three benchmarks
(Appendix~\ref{sec:retrieval_composition}), confirming that PPR
activates every retrievable node type rather than collapsing to a
text-only retriever; a retrieval-only Recall@K comparison against
the IR baseline is in Appendix~\ref{sec:recall_at_k}. The full
subtask-by-subtask analysis is in
Appendix~\ref{sec:mmg_mechanism}.

\textbf{Case study.} A representative round-by-round trace illustrating MMG's
cross-modal advantage on EgoLifeQA Q201 is given in
Appendix~\ref{sec:case_study_appendix} (Case~1).

\subsection{Impact of the Narrative Memory Chain}
\label{sec:nmc_analysis}

We now drill into \emph{where} NMC's $+3.2$ / $+3.4$ / $+3.1$-point
lift on EgoLifeQA / Ego-R1 / MM-Lifelong concentrates, by stratifying
every question by which chain type (if any) fires on it at
retrieval time (Table~\ref{tab:coverage}).

\begin{table}[t]
\caption{Per-question A/B comparison of MAGIC-Video's full system
(\emph{+ NMC}) against its graph-only variant (\emph{+ MMG}, no
chain injection), stratified by whether any chain fires on the
question.}
\label{tab:coverage}
\centering
\resizebox{\textwidth}{!}{%
\begin{tabular}{l ccc c ccc c ccc}
\toprule
\multirow{2}{*}{Subset}
& \multicolumn{3}{c}{EgoLifeQA (500)}
& & \multicolumn{3}{c}{Ego-R1 (150, 3 runs)}
& & \multicolumn{3}{c}{MM-Lifelong (623)} \\
\cmidrule(lr){2-4} \cmidrule(lr){6-8} \cmidrule(lr){10-12}
& n & + MMG & + NMC ($\Delta$)
& & n & + MMG & + NMC ($\Delta$)
& & n & + MMG & + NMC ($\Delta$) \\
\midrule
All            & 500 & 64.4 & 67.6 {\scriptsize ($+$3.2)}
& & 150 & 61.3 & 64.7 {\scriptsize ($+$3.4)}
& & 623 & 21.4 & 24.5 {\scriptsize ($+$3.1)} \\
\midrule
None           & 268 & 63.4 & 64.6 {\scriptsize ($+$1.2)}
& & 69  & 62.3 & 59.4 {\scriptsize ($-$2.9)}
& & 305 & 22.6 & 26.9 {\scriptsize ($+$4.3)} \\
Topic only     & 166 & 66.9 & 70.5 {\scriptsize ($+$3.6)}
& & 56  & 57.1 & 62.5 {\scriptsize ($+$5.4)}
& & 126 & 23.0 & 28.2 {\scriptsize ($+$5.2)} \\
Event only &  21 & 61.9 & 71.4 {\scriptsize ($+$9.5)}
& &  5  & 100.0 & 100.0 {\scriptsize ($+$0.0)}
& & 102 & 19.6 & 22.5 {\scriptsize ($+$2.9)} \\
Both           &  45 & 62.2 & 73.3 {\scriptsize ($+$11.1)}
& & 20  & 60.0 & 80.0 {\scriptsize ($+$20.0)}
& &  90 & 16.7 & 13.3 {\scriptsize ($-$3.4)} \\
Any            & 232 & 65.5 & 71.1 {\scriptsize ($+$5.6)}
& & 81  & 60.5 & 69.1 {\scriptsize ($+$8.6)}
& & 318 & 20.1 & 22.2 {\scriptsize ($+$2.1)} \\
\bottomrule
\end{tabular}%
}
\end{table}

\textbf{Mechanism.} NMC fires on $46$--$54\%$ of questions across
the three benchmarks, with topic chains firing several times more
often than event chains. On the egocentric benchmarks the
\emph{Any} (chain-firing) lift dwarfs the \emph{None} cell
($+5.6$ vs.\ $+1.2$ on EgoLifeQA; $+8.6$ vs.\ $-2.9$ on Ego-R1),
pinning the gain to the chain-firing population. Topic-only and
Event-only each contribute positively in isolation, and the
\emph{Both} subset records the largest single controlled gain
within its benchmark column on EgoLifeQA ($+11.1$, $n{=}45$) and
Ego-R1 ($+20.0$, $n{=}20$). MM-Lifelong shows the same direction: every chain-firing cell is
positive (Topic-only $+5.2$, Event-only $+2.9$, Any $+2.1$),
confirming that NMC contributes a consistent positive lift on the
open-ended setting as well; the small negative in the Both cell
likely reflects the doubled chain context diluting the query signal
on open-ended low-baseline questions. Full per-cell analysis
is in Appendix~\ref{sec:nmc_mechanism}.

\textbf{Case studies.} Representative round-by-round traces of NMC's chain-injection
advantage (Case~2) and its context-dilution failure on the Both cell
(Case~3) are in Appendix~\ref{sec:case_study_appendix}.

\subsection{Efficiency Analysis}
\label{sec:efficiency}

Having attributed the accuracy gains to MMG and NMC, we ask
whether they also raise the inference cost. We answer along two
axes: a per-stage \emph{complexity analysis} that counts the
operations by each pipeline stage
(Table~\ref{tab:efficiency}), and a \emph{search-round analysis}
that compares MAGIC-Video's online agentic budget against the
independent-retrieval baseline (Table~\ref{tab:rounds}).

\textbf{Complexity analysis.} \emph{Offline.} Preprocessing and MMG
construction together issue $O(N)$ LLM calls and $O(N)$ encoder
forward passes, with $N$ the number of captioned video hours; NMC
distillation adds an $E + D + 2V + 4C$ term that stays a small
fraction of preprocessing even at month-long scales. All offline
artefacts are built once per corpus and amortize across every
question asked of it. \emph{Online.} The agentic loop is capped at
$5$ search rounds and consumes on average $\sim\!4.4$ controller
LLM calls, $\sim\!3.3$ PPR + NMC lookups each, $\sim\!7$ embedding
forwards, and one final answer call per question, comparable to
agentic peers such as EGAgent and WorldMM. Per-stage breakdowns
and full formulas are in Appendix~\ref{sec:efficiency_full}.

\begin{table}[t]
\centering
\small
\caption{Per-stage operation count. $N$ = hours of captioned video,
$V$ = video sources, $D$ = days, $E$ = unique entities, $C$ = event
chains. A full version of per-stage breakdown is given in
Appendix~\ref{sec:efficiency_full}.}
\label{tab:efficiency}
\begin{tabular}{llr}
\toprule
Stage & Op type & Count (formula) \\
\midrule
\rowcolor{gray!12} \multicolumn{3}{l}{\emph{Offline (built once per corpus)}} \\
\makecell[l]{Preprocessing (Sec.~\ref{sec:preprocessing}) \\ + MMG construction( Sec.~\ref{sec:mmg})} & GPU/API & $\sim\!4500\,N$ \\
NMC distillation (Sec.~\ref{sec:nmc})        & GPU/API & $E + D + 2V + 4C$ \\
\midrule
\rowcolor{gray!12} \multicolumn{3}{l}{\emph{Online (per question, Sec.~\ref{sec:agentic_retrieval})}} \\
Query text+visual embedding  & GPU/API & $\le\!5$ \\
Cross-modal PPR + NMC lookup & CPU     & $\le\!5$ \\
Retrieval controller         & GPU/API & $\le\!5$ \\
Answer backbone              & GPU/API & 1 \\
\bottomrule
\end{tabular}
\end{table}

\textbf{Search-round and retrieval-token analysis.} We further verify
that MAGIC-Video does not inflate the
agentic search budget relative to the WorldMM-style
independent-retrieval baseline.
Table~\ref{tab:rounds} reports the mean number of retrieval rounds,
the share of queries that exhaust the 5-round cap, and the mean
retrieval-context tokens fed to the controller per question on
EgoLifeQA, where a matched IR run is available. MAGIC-Video uses
slightly \emph{fewer} rounds than IR ($3.33$ vs $3.49$) and a
lower @5-cap rate ($29.4\%$ vs $32.2\%$), so pre-distilled chains
let the agent reach an answer without spending more agentic turns.
The per-question retrieval context is $\sim\!1.78\times$ larger in
text tokens and includes $\sim\!3.75\times$ more visual frames
($51.0$ vs $13.6$); both gaps reflect cross-modal PPR's visual
coverage on essentially every query ($98\%$ vs IR's $15\%$,
Appendix~\ref{sec:retrieval_composition}) under WorldMM's
single-modality routing, which directs only $\sim\!8\%$ of rounds
to the visual index. This is the visual evidence behind the
$+12.0/+11.1$\,pp lifts on EntityLog/EventRecall
(Sec.~\ref{sec:mmg_analysis}) and the direct price of the $+11.6$
accuracy gain over IR ($56.0 \to 67.6$).

\begin{table}[t]
\centering
\small
\caption{Search-round and retrieval-content stats on EgoLifeQA
(IR vs.\ MAGIC-Video). ``Mean rd.'' is mean retrieval rounds per
question; ``@5-cap'' is the share hitting the 5-round budget;
``Text tok/q'' is the mean retrieval-content text tokens fed to
the controller. Image-token cost is model-dependent, so we report
``Frames/q'' (mean retrieved frames per question) separately.}
\label{tab:rounds}
\begin{tabular}{lrrrr}
\toprule
Config & Mean rd. & @5-cap & Text tok/q & Frames/q \\
\midrule
IR          & 3.49 & 32.2\% & 3{,}471 & 13.6 \\
MAGIC-Video & 3.33 & 29.4\% & 6{,}189 & 51.0 \\
\bottomrule
\end{tabular}
\vspace{-0.3cm}
\end{table}

\section{Conclusion}
\label{sec:conclusion}

We presented MAGIC-Video, a training-free \emph{structured-memory}
framework for ultra-long video reasoning that pairs a multimodal
memory graph with a narrative memory chain. The graph unifies episodic, semantic, and visual evidence
under six typed edges so that a single cross-modal Personalized
PageRank pass propagates relevance across modality boundaries; the
chain distils long-horizon entity biographies and recurring activity
events that no single retrieved snippet can express. At inference time, an agentic loop combines per-round graph
retrieval with narrative chain injection and feeds the merged context to a reasoning backbone. Across
EgoLifeQA, Ego-R1, and MM-Lifelong, MAGIC-Video consistently
outperforms strong general-purpose, long-video, RAG-based, and agentic
baselines, with the largest gains on subtasks that require cross-modal
or cross-day reasoning.

\section{Limitations and future work}
\label{sec:limitations}

Despite MAGIC-Video's strong performance across the three benchmarks,
the framework has three limitations that each suggest a natural future
direction.
\textbf{(i) Caption-bounded recall.} Both retrieval and narrative
distillation are upper-bounded by the raw captions: an event that is not
described in any caption cannot be surfaced by graph traversal or chain
matching. \emph{Future work:} pair the pipeline with a domain-adapted
captioner, or invoke an external visual tool on demand when the
controller signals that retrieved captions are insufficient.
\textbf{(ii) Offline construction.} Both MMG and NMC are built once,
offline, so streaming video, where captions, entities, and narratives
must be updated as new footage arrives, is not yet supported.
\emph{Future work:} an online variant that refreshes the graph and chains
incrementally per 30-s window, with bounded re-ranking on the affected
subgraph rather than rebuilding from scratch.
\textbf{(iii) Higher per-question token consumption.} Cross-modal PPR
returns a multi-scale node mix and NMC injects additional chain facts
when a chain matches the query, which on EgoLifeQA raises the
per-question retrieval context from $\sim\!3.5$\,K to $\sim\!6.2$\,K tokens
($\sim\!1.78\times$ the IR baseline) for an $+11.6$\,pp accuracy gain
(Sec.~\ref{sec:efficiency}); the round count itself does not increase,
but the controller's prompt grows accordingly. \emph{Future work:} a
relevance-pruning pass after each search round that learns which
retrieved items the reasoning backbone actually uses, and drops the rest
before re-feeding the merged context.

\newpage

\bibliographystyle{plainnat}
\bibliography{refs}

%%%%%%%%%%%%%%%%%%%%%%%%%%%%%%%%%%%%%%%%%%%%%%%%%%%%%%%%%%%%
\newpage
\appendix

\begin{center}
\Large\textbf{Appendix --- Table of Contents}
\end{center}
\vspace{2em}

\noindent
\begin{itemize}
\setlength\itemsep{0.5em}
\item[\textbf{A}\hspace{0.5em}] Extended related work \dotfill p.\ \pageref{sec:related_extended}
\item[\textbf{B}\hspace{0.5em}] Broader impacts \dotfill p.\ \pageref{sec:broader_impacts}
\item[\textbf{C}\hspace{0.5em}] Multimodal memory graph: details \dotfill p.\ \pageref{sec:mmg_details}
\item[\textbf{D}\hspace{0.5em}] Narrative memory chain: details \dotfill p.\ \pageref{sec:nmc_details}
\item[\textbf{E}\hspace{0.5em}] Models and preprocessing details \dotfill p.\ \pageref{sec:impl_details}
\item[\textbf{F}\hspace{0.5em}] Hyperparameters and controlled-comparison protocol \dotfill p.\ \pageref{sec:hyperparams}
\item[\textbf{G}\hspace{0.5em}] Retrieval composition by node type \dotfill p.\ \pageref{sec:retrieval_composition}
\item[\textbf{H}\hspace{0.5em}] Retrieval Recall@K analysis \dotfill p.\ \pageref{sec:recall_at_k}
\item[\textbf{I}\hspace{0.5em}] Full per-stage cost breakdown \dotfill p.\ \pageref{sec:efficiency_full}
\item[\textbf{J}\hspace{0.5em}] Statistical significance of the headline gains \dotfill p.\ \pageref{sec:significance}
\item[\textbf{K}\hspace{0.5em}] Judge robustness on MM-Lifelong \dotfill p.\ \pageref{sec:judge_robustness}
\item[\textbf{L}\hspace{0.5em}] Qualitative case studies \dotfill p.\ \pageref{sec:case_study_appendix}
\item[\textbf{M}\hspace{0.5em}] Prompt templates \dotfill p.\ \pageref{sec:prompts}
\end{itemize}

\newpage
\section{Extended related work}
\label{sec:related_extended}

\textbf{Long Video Understanding.}
General-purpose multimodal foundation models have pushed context windows to
the million-token scale, with frontier systems such as
Gemini~3.1~\citep{gemini31} and GPT-5.4~\citep{gpt54} reporting up to 1M-token contexts.
Video-specific models, including LLaVA-Video~\citep{llavavideo},
VideoChat-Flash~\citep{videochatflash}, Time-R1~\citep{timer1}, and
Video-RTS~\citep{videorts}, further optimise temporal modelling through token
compression, hierarchical sampling, or reinforcement learning. Despite these
advances, even a million-token budget covers only tens of minutes of densely
sampled video, far short of the days- or weeks-long recordings found in
egocentric life logging, livestream recordings, or surveillance archives. At such scales, any
method that tries to fit raw video into the context window is forced into
aggressive downsampling, inevitably discarding query-relevant details that
appear briefly or are scattered across long horizons.

\textbf{Agentic AI and LLM-based Reasoning Agents.}
A broader strand of work casts LLMs as autonomous reasoning agents that
interleave \emph{thinking}, \emph{acting}, and \emph{remembering} across
multiple decision rounds rather than answering in a single forward pass.
ReAct~\citep{react} introduced the foundational pattern of alternating
natural-language reasoning with discrete actions; Reflexion~\citep{reflexion}
adds verbal self-evaluation across rounds so agents can recover from earlier
mistakes; and Tree-of-Thoughts~\citep{tot} generalises the action stream to
a deliberate search over reasoning branches. A complementary line equips
agents with external tools: Toolformer~\citep{toolformer} fine-tunes models
to invoke APIs in-line, and ToolLLM~\citep{toolllm} scales tool repertoires
to thousands of real-world endpoints. Retrieval-focused agents fold these
primitives into the RAG loop --- Self-RAG~\citep{selfrag} learns when to
retrieve and when to stop via reflection tokens, and
Search-o1~\citep{searcho1} interleaves retrieval calls inside a long chain
of thought. Closest in spirit to our setting are agentic memory
architectures such as Generative Agents~\citep{generativeagents}, which
equip long-horizon agents with structured event-stream memory that persists
across episodes. A range of concurrent work continues to expand the
agentic-AI design space along routing, multi-agent collaboration, and
graph-structured perspectives~\citep{agentrouter,evolverouter,autodata,graphsubstrate,mass}.
MAGIC-Video can be viewed as an instance of the same paradigm, specialised
to the ultra-long video regime: the controller emits
\texttt{search}/\texttt{answer} decisions in the ReAct style, but each
\texttt{search} step is a single cross-modal PPR pass over the
offline-distilled MMG plus a matching step on NMC, rather than a free-form
tool call --- so the agentic loop and the structured memory are co-designed
to amortise offline distillation cost across all queries on the same video.

\textbf{Memory-Augmented and Agentic Video Reasoning.}
To move beyond the context window, recent work equips video models with
external memory or multi-step reasoning. One line uses unstructured or lightly
structured memory: VideoMem~\citep{videomem} and
EventMemAgent~\citep{eventmemagent} maintain compact textual or event-tuple
memory updated by RL-trained policies, while
LongVideoAgent~\citep{longvideoagent} coordinates sub-agents that issue
grounding and visual queries. A complementary line builds structured graphs as
memory. EGAgent~\citep{egagent} stores entities in an SQL-queryable scene
graph and dispatches visual, audio, and graph searches as separate tools;
M3-Agent~\citep{m3agent} builds an entity-centric graph whose nodes hold
textual, visual, or audio content linked predominantly by cross-modal
entity-identity edges;
MM-Mem~\citep{mmmem} organises memory as a three-level pyramid with a
symbolic graph only at the top layer; AVI~\citep{avi} couples a text-only
temporal entity graph with on-demand perception tools; and Vgent~\citep{vgent}
and VideoRAG~\citep{videorag} combine graph-structured textual knowledge with
dual-channel multimodal retrieval. The concurrent VimRAG~\citep{vimrag}
builds a dynamic DAG \emph{over the agent's reasoning trajectory}, whose
nodes record per-step retrieved visual tokens rather than the video's own
content structure; and WorldMM~\citep{worldmm} keeps three heterogeneous
memories---a multi-scale textual event graph, a semantic knowledge graph,
and a hybrid embedding-plus-frame visual store---and relies on a single
adaptive agent that iteratively selects among them per query. Collectively, even the systems that place multimodal content on
graph nodes (e.g., M3-Agent's per-entity text/visual/audio attachments)
retrieve through per-modality indices or per-tool channels orchestrated
by the agent, rather than a single graph-native pass that propagates
relevance across modalities. We argue instead that the memory itself
should be multimodal and that reasoning over it should be graph-native.
We therefore propose a unified content graph in which episodic, semantic, and
visual evidence co-exist as first-class nodes, and a cross-modal retrieval
mechanism that propagates query relevance through the graph in a single pass.

\textbf{Graph-Based Knowledge Retrieval.}
Graph-based retrieval has become a prominent alternative to dense-vector
search in RAG systems. GraphRAG~\citep{graphrag} constructs a knowledge graph
from documents and retrieves via hierarchical community summaries;
LightRAG~\citep{lightrag} simplifies graph construction and introduces
dual-level (local and global) retrieval; HippoRAG~\citep{hipporag} runs
Personalized PageRank (PPR) over an entity graph to surface multi-hop
evidence more reliably than similarity search. Subsequent work extends this
paradigm along several axes: Zep~\citep{zep} and MemoTime~\citep{memotime}
incorporate bi-temporal modelling for reasoning about when facts hold;
MAGMA~\citep{magma} explores multi-view graphs that separate semantic,
temporal, and causal relations; and G-Memory~\citep{gmemory} and
Optimus-1~\citep{optimus} introduce hierarchical graphs for agent memory.
All of these operate over purely textual nodes, leaving open the question of
how graph-based retrieval can be applied to inherently multimodal
data---a gap we address by admitting visual clips as first-class nodes within
the retrieval graph.

\newpage
\section{Broader impacts}
\label{sec:broader_impacts}

MAGIC-Video targets question answering over ultra-long egocentric and
livestream video. Several positive applications follow naturally from
this setting: (i) long-horizon \emph{memory assistants} for users
with cognitive impairments or for elderly users who benefit from
being able to query their own life-log on demand; (ii) \emph{archive
analysis} for researchers, journalists, and historians who need to
trace recurring entities or multi-day events across long footage
without watching it linearly; and (iii) \emph{personal productivity}
applications such as automated meeting recall or activity-based
journaling.

The same retrieval capability also raises potential negative
impacts that we want to acknowledge explicitly. \emph{Surveillance
over-reach}: a system that can answer "when did X first appear?" or
"how did Y unfold across the week?" lowers the technical barrier to
person-tracking analysis on continuous video, which can be misused
on non-consenting subjects. \emph{Privacy in life-logging}: any
agent that records day-to-week-long egocentric video records its
wearer's housemates, colleagues, and bystanders, who may not have
consented to their utterances or actions being indexed and queried.
\emph{Bias in distillation}: the LLMs used for caption merge, NER,
and chain distillation can carry social or demographic biases that
propagate into the entity biographies and event chains
MAGIC-Video produces; downstream applications that rely on these
artefacts inherit those biases.

We do not introduce a new generative capability beyond what existing
video-LLMs and graph-RAG systems already enable, and our reported
experiments use only public benchmarks
(EgoLifeQA, Ego-R1, MM-Lifelong) with their standard evaluation
splits. Mitigations for the concerns above include (a) gating
deployment of life-log retrieval to consented participants,
(b) preserving deletion rights at the entity / chain level so an
individual can request removal of their footprint from the
distilled artefacts, and (c) auditing the underlying LLM's biases
before applying MAGIC-Video to high-stakes downstream decisions.

\newpage
\section{Multimodal memory graph: details}
\label{sec:mmg_details}

This appendix gives the implementation-level details of the
Multimodal Memory Graph: aggregate sizes
(Sec.~\ref{sec:graph_stats}), the per-edge-type weight schedule that
defines PPR's transition matrix (Sec.~\ref{sec:edge_weights}), and a
worked instantiation of all four node types and all six edge types
on a single 30-second window from EgoLifeQA's A1\_JAKE recording
(Sec.~\ref{sec:mmg_example}).

\subsection{Aggregate statistics}
\label{sec:graph_stats}

Table~\ref{tab:graph_stats} reports the active-subgraph statistics
for both ultra-long video corpora on which MAGIC-Video is evaluated:
EgoLifeQA's A1\_JAKE recording (one continuous 51.9-hour, 7-day
first-person footage) and the MM-Lifelong Month split (14 livestream
videos totalling 76.4 captioned hours; the 14 sources cover the 623
evaluation questions). All numbers are the sizes of the time-filtered
subgraph $\mathcal{G}_{\le t}$ on which PPR actually runs, after
\texttt{filter\_by\_time} collapses the cumulative consolidation
snapshots in \texttt{graph.pkl} to the latest snapshot $\le t$ (the
on-disk pre-collapse counts are listed in the table caption for
completeness).

\begin{table}[h]
\caption{MAGIC-Video active subgraph sizes after
\texttt{filter\_by\_time}, the granularity at which PPR runs.
Numbers are exact: for EgoLifeQA we filter the single A1\_JAKE
\texttt{graph.pkl}; for MM-Lifelong Month we filter and sum across
all 14 per-broadcast graphs. The offline serialised \texttt{graph.pkl}
stores cumulative consolidation snapshots (1{,}330{,}179 / 54{,}375
semantic-triple rows on disk for the two benchmarks); on EgoLifeQA
this collapses dramatically because the week-long recording produces
many cumulative snapshots, whereas on MM-Lifelong each broadcast's
disk count already approximates its active count.
\texttt{has\_property} is the only edge type whose count differs
between the on-disk graph and the active subgraph; the other five
types are unaffected. Pkl size is the on-disk footprint of the
serialised graph object.}
\label{tab:graph_stats}
\centering
\begin{tabular}{lrr}
\toprule
 & EgoLifeQA A1\_JAKE & MM-Lifelong Month \\
\midrule
\textbf{Total active nodes} & \textbf{20{,}338} & \textbf{94{,}857} \\
\quad Episode (all granularities) & 7{,}625 & 11{,}311 \\
\quad\quad 30\,s & 6{,}223 & 9{,}168 \\
\quad\quad 3\,min & 1{,}069 & 1{,}535 \\
\quad\quad 10\,min & 269 & 517 \\
\quad\quad 1\,h & 64 & 91 \\
\quad Visual-clip & 6{,}223 & 9{,}168 \\
\quad Entity & 2{,}669 & 23{,}210 \\
\quad Semantic-triple (latest snapshot per query) & 3{,}821 & 51{,}168 \\
\midrule
\textbf{Total active edges} & \textbf{82{,}446} & \textbf{244{,}014} \\
\quad \texttt{has\_property} (active per query) & 4{,}020 & 77{,}690 \\
\quad \texttt{mentioned\_in} & 18{,}619 & 52{,}552 \\
\quad \texttt{appeared\_in} (shortcut) & 18{,}619 & 52{,}538 \\
\quad \texttt{temporal\_next} & 15{,}182 & 22{,}510 \\
\quad \texttt{contains} & 13{,}560 & 20{,}388 \\
\quad \texttt{co\_clip} & 12{,}446 & 18{,}336 \\
\midrule
Pkl size on disk & 503\,MB & 523\,MB \\
\bottomrule
\end{tabular}
\end{table}

\subsection{Edge weights}
\label{sec:edge_weights}

All edges enter the PPR transition matrix $\tilde{A}$ with explicit
positive weights; nodes themselves carry no per-node weight, and the
only node-level personalisation comes from the query-dependent reset
vector~$\mathbf{r}$ (Eq.~\ref{eq:reset}). The full weight schedule
is given in Table~\ref{tab:edge_weights}: structural same-modality /
same-window edges (\textsc{co\_clip}, \textsc{mentioned\_in},
adjacent-layer \textsc{contains}, subject-side \textsc{has\_property})
get the default weight $1.0$; \emph{derived} or \emph{weak-evidence}
edges are down-weighted to $0.5$--$0.8$ so that PPR mass leaks across
them more conservatively; and \textsc{temporal\_next} gets a continuous
time-decay $1/(\Delta t_{\text{sec}} + 1)$ on the gap between
\emph{adjacent same-granularity} windows: in continuous recording the
gap is $0$ and the edge weight is $\approx\!1$, while any non-trivial
pause attenuates it sharply (a 1-min pause already drops $w$ to
$\sim\!0.02$); edges with gaps beyond $6$\,h are dropped entirely.

\begin{table}[h]
\centering\small
\caption{Edge-weight schedule used by MMG. PPR is run with
\texttt{directed=False}, so all edges act as bidirectional channels
for relevance flow regardless of how they are stored.}
\label{tab:edge_weights}
\begin{tabular}{@{}lll@{}}
\toprule
\textbf{Edge type} & \textbf{Weight} & \textbf{Comment} \\
\midrule
\textsc{co\_clip}                      & $1.0$ & same time window \\
\textsc{mentioned\_in}                 & $1.0$ & entity named in 30-s caption \\
\textsc{has\_property} (subject)       & $1.0$ & entity is the triple's subject \\
\textsc{has\_property} (object)        & $0.5$ & weaker than subject role \\
\textsc{appeared\_in} (shortcut)       & $0.8$ & derived via \textsc{mentioned\_in}\,$\to$\,\textsc{co\_clip} \\
\textsc{contains} (adjacent layer) & $1.0$ & 3\,min\,$\to$\,30\,s; 10\,min\,$\to$\,3\,min; 1\,h\,$\to$\,10\,min \\
\textsc{contains} (skip layer)         & $0.5$ & 1\,h\,$\to$\,30\,s direct shortcut \\
\textsc{temporal\_next}                & $1/(\Delta t_{\text{sec}}+1)$ & same granularity, adjacent in time \\
\bottomrule
\end{tabular}
\end{table}

\subsection{Worked example: a 30-second window in MMG}
\label{sec:mmg_example}

To make the schema concrete, we instantiate every node and edge type
on a single 30-second window from EgoLifeQA's A1\_JAKE recording
(DAY1 17:42:01--17:42:30, from the Day-1 supermarket scene from which
Case Study~1 in Appendix~\ref{sec:case_study_appendix} is drawn).
All examples below are taken verbatim from the constructed
\texttt{graph.pkl}.

\paragraph{Node-type instantiations.}
\begin{itemize}\setlength\itemsep{0.2em}
\item \emph{Episode (30\,s)}: caption ---
\begin{quote}
\colorbox{black!5}{%
\parbox{0.80\linewidth}{\small \texttt{[DAY1 17:42:01--17:42:30]}
``I stop and look at everyone's shopping carts. Katrina says,
`Done shopping,' then notes, `There are two missing.' I reply,
`You guys\dots' We chat for a while. Alice mentions, `Seasonings.'
I ask, `Did you buy the meat?' Katrina asks, `So, we're buying
hot pot ingredients, right?'\dots''}}
\end{quote}

\item \emph{Episode (3\,min)}: LLM-aggregated summary that covers
the same window plus its two neighbours ---
\begin{quote}
\colorbox{black!5}{%
\parbox{0.80\linewidth}{\small \texttt{[DAY1 17:42:00--17:45:00]}
``I and Katrina compare carts and discover items are missing\dots\
Shure suggests buying hot-pot ingredients first and mentions order,
delivery options, and fridge capacity, while the group moves on
with Shure, Tasha, Alice, and Lucia.''}}
\end{quote}

\item \emph{Episode (10\,min and 1\,h)}: produced by the same
recursive aggregation applied to coarser windows. Their node ids
are
\begin{quote}
\colorbox{black!5}{%
\parbox{0.80\linewidth}{\small\ttfamily
episode\_10min\_DAY1\_17:40--17:50\\
episode\_1h\_DAY1\_17:00--18:00}}
\end{quote}
each storing an LLM-summarised caption over its interval.

\item \emph{Visual-clip}: a 3584-d visual embedding of 16~uniformly
sampled frames covering the 30-s window; the text-embedding field
is zeroed (Sec.~\ref{sec:mmg}). Node id:
\begin{quote}
\colorbox{black!5}{%
\parbox{0.80\linewidth}{\small\ttfamily visual\_clip\_DAY1\_17:42:00--17:42:30}}
\end{quote}

\item \emph{Entity}: a canonical entity produced by NER, with the
text-embedding field also zeroed. Node id:
\begin{quote}
\colorbox{black!5}{%
\parbox{0.80\linewidth}{\small\ttfamily entity\_Katrina}}
\end{quote}

\item \emph{Semantic}: a consolidated $(s,p,o)$ tuple extracted from
this window, with id of form \texttt{triple\_<snapshot\_ts>\_<idx>}
so that distinct extraction events at different timestamps remain
distinct nodes (no $(s,p,o)$-level deduplication). Node id and
content for this window:
\begin{quote}
\colorbox{black!5}{%
\parbox{0.80\linewidth}{\small\ttfamily triple\_DAY1\_17:42:01\_03 \\
\quad= (Katrina,\ asks\ about,\ hot\ pot\ ingredients)}}
\end{quote}
\end{itemize}

\paragraph{Edge-type instantiations.}
\begin{itemize}\setlength\itemsep{0.2em}
\item \textsc{mentioned\_in}: the entity name appears verbatim in
the 30-s caption.
\begin{quote}
\colorbox{black!5}{%
\parbox{0.80\linewidth}{\small\ttfamily entity\_Katrina\;$\rightarrow$\;episode\_30s\_DAY1\_17:42:01}}
\end{quote}

\item \textsc{co\_clip}: same 30-s time window.
\begin{quote}
\colorbox{black!5}{%
\parbox{0.80\linewidth}{\small\ttfamily episode\_30s\_DAY1\_17:42:01\;\textendash\;visual\_clip\_DAY1\_17:42:00}}
\end{quote}

\item \textsc{appeared\_in}: shortcut materialised whenever a
\textsc{mentioned\_in}\,$\to$\,\textsc{co\_clip} path exists.
\begin{quote}
\colorbox{black!5}{%
\parbox{0.80\linewidth}{\small\ttfamily entity\_Katrina\;$\rightarrow$\;visual\_clip\_DAY1\_17:42:00}}
\end{quote}

\item \textsc{has\_property}: entity is the subject of the triple.
\begin{quote}
\colorbox{black!5}{%
\parbox{0.80\linewidth}{\small\ttfamily entity\_Katrina\;$\rightarrow$\;triple\_DAY1\_17:42:01\_03}}
\end{quote}

\item \textsc{temporal\_next}: adjacent same-granularity episodes,
weight $w \propto 1/\Delta t$.
\begin{quote}
\colorbox{black!5}{%
\parbox{0.80\linewidth}{\small\ttfamily episode\_30s\_DAY1\_17:42:01\;\textendash\;episode\_30s\_DAY1\_17:42:31}}
\end{quote}

\item \textsc{contains}: the 30-s interval falls inside the 3-min one.
\begin{quote}
\colorbox{black!5}{%
\parbox{0.80\linewidth}{\small\ttfamily episode\_3min\_DAY1\_17:42:00\;$\rightarrow$\;episode\_30s\_DAY1\_17:42:01}}
\end{quote}
\end{itemize}

This single window therefore participates in 6 of the 6 edge types
and contributes to 4 of the 4 node types. The full A1\_JAKE graph
contains $6{,}223$ such 30-s windows; aggregate counts per node and
edge type appear in Table~\ref{tab:graph_stats} and edge weights in
Table~\ref{tab:edge_weights}.

\subsection{MMG mechanism: subtask-by-subtask analysis}
\label{sec:mmg_mechanism}

This appendix expands Sec.~\ref{sec:mmg_analysis} with the
typed-edge mechanism behind each subtask gain. EntityLog questions
(e.g., \emph{``who used the screwdriver first?''}) require resolving
a named entity to the exact moment when it first appears, and
EventRecall questions (e.g., \emph{``when did X happen?''}) require
attaching a textual cue to a timestamped visual clip. Both correspond
to the typed-edge path MMG is designed to support: a text or BM25
seed activates an Entity node via \textsc{mentioned\_in}, which in
turn activates the co-clipped visual clip via the \textsc{co\_clip}
edge, so a single cross-modal PPR pass co-retrieves the textual
mention and the visual moment without an explicit modality switch
in the agent loop. The smaller RelationMap lift ($+3.2$\,pp) reflects
a ceiling effect: relational queries between persons are typically
recoverable from textual context alone (the IR baseline already
reaches $62.4\%$), so MMG inherits rather than amplifies that
performance. Together this confirms that the graph delivers its
lift precisely where the typed cross-modal edges are the enabling
machinery, not as a uniform ``bigger retrieval budget''
improvement.

\newpage
\section{Narrative memory chain: details}
\label{sec:nmc_details}

This appendix gives the implementation-level details of the
Narrative Memory Chain: aggregate sizes of the offline-distilled
chain set (Sec.~\ref{sec:chain_stats}), the online injection
procedure that admits chains into a round's retrieved set
(Sec.~\ref{sec:nmc_injection}), and a per-cell A/B mechanism
analysis of where NMC's accuracy lift comes from
(Sec.~\ref{sec:nmc_mechanism}).

\subsection{Aggregate statistics}
\label{sec:chain_stats}

Table~\ref{tab:chain_stats} reports the size of the Narrative Memory
Chain artefacts produced by the offline distillation pipeline of
\S\ref{sec:nmc} for the same two ultra-long video corpora as the
graph statistics: EgoLifeQA's A1\_JAKE recording, which is also the
source video used by the Ego-R1 evaluation slice, and the
MM-Lifelong Month split (14 livestream videos aggregated). For
\emph{topic chains} we report the entities that yield at least one
extracted dated fact after the LLM distillation pass, i.e.\ those
that the candidate-selection stage marks as having a non-trivial
biographic lifecycle; entities for which the distiller returns an
empty fact list ($\sim\!8\%$ of NER-extracted entities on EgoLifeQA
and $\sim\!6\%$ on MM-Lifelong, typically one-off locations, brand
names, or abstract concepts) are excluded from the chain count
because they contribute nothing to NMC injection at query time. For
\emph{event chains} we report all three distillation stages: the
per-day activity log produced at Step~1, the cross-time chains
discovered at Step~2, and the final enriched chain set produced at
Step~3, where each chain is expanded into a list of dated steps. The
two corpora differ in granularity: A1\_JAKE is a single 7-day
continuous recording, so Step~1 produces one activity log per day
(7 days, 89 activities), and Step~2 discovers a small number of
high-coverage multi-day arcs (18 chains, 75 enriched steps).
MM-Lifelong runs the same pipeline independently per video source,
so the totals add up across all 14 videos, with one-day activity
logs per video (14 days, 785 activities) and 173 chains carrying
775 enriched steps. The fact that EgoLifeQA produces only
$\sim\!10\!\times$ fewer event chains than MM-Lifelong despite
covering only $1/14$th of the independent video sources reflects
the longer cross-day events that A1\_JAKE supports as a single
continuous 7-day recording.

\begin{table}[h]
\caption{MAGIC-Video Narrative Memory Chain sizes. Topic-chain rows
count entities that yield at least one dated fact after the
distillation pass. Event-chain rows trace the three NMC distillation
stages of \S\ref{sec:nmc}: Step~1 per-day activity extraction,
Step~2 cross-time chain discovery, and Step~3 step enrichment.
Ego-R1 is omitted as a separate column because its evaluation slice
draws from the same A1\_JAKE recording as EgoLifeQA and therefore
inherits the same NMC artefacts.}
\label{tab:chain_stats}
\centering
\resizebox{\textwidth}{!}{%
\begin{tabular}{lrr}
\toprule
 & EgoLifeQA / Ego-R1 (A1\_JAKE) & MM-Lifelong Month \\
\midrule
\rowcolor{gray!10} \multicolumn{3}{l}{\emph{Topic chains (per-entity biographies)}} \\
\quad Entity biographies (with $\ge 1$ dated fact) & 236 & 1{,}574 \\
\quad Dated facts (total) & 5{,}267 & 10{,}359 \\
\midrule
\rowcolor{gray!10} \multicolumn{3}{l}{\emph{Event chains (multi-day activities)}} \\
\quad Step~1 (days $\times$ activities) & 7 days, 89 activities & 14 days, 785 activities \\
\quad Step~2 chains discovered & 18 & 173 \\
\quad Step~3 enriched chains & 18 & 173 \\
\quad Step~3 steps (total) & 75 & 775 \\
\bottomrule
\end{tabular}%
}
\end{table}

\subsection{Online NMC injection procedure}
\label{sec:nmc_injection}

At each search round, after PPR retrieval, NMC injection runs
independently for topic chains and event chains and produces the
round's $\mathcal{N}_r$ in Eq.~\ref{eq:context}.
\textbf{Candidate filter.} A chain is a candidate only if it is
\emph{anchored} by this round's retrieval: a topic chain qualifies
when its entity name (case-/plural-insensitive regex) appears in
$\geq \texttt{min\_hits}$ retrieved 30-s captions; an event chain
qualifies when $\geq \texttt{storyline\_min\_hits}$ of its step
time ranges contain a retrieved episode.
\textbf{Tier 1 (keyword).} Candidates whose keyword entities
(entity name for topic chains; \texttt{key\_entities} for event
chains) match $q$ via the same regex are admitted up to a per-type
slot budget.
\textbf{Tier 2 (embedding).} Remaining candidates fill the rest of
the budget by cosine similarity between $q$ and a concatenation of
the chain's content (facts for topic chains; step descriptions for
event chains), keeping only chains above a similarity threshold,
ranked by cosine.
\textbf{Post-filtering.} Admitted facts are time-filtered by $t$
(drop facts with $t_i > t$) and deduplicated against any retrieved
episode whose interval ($\pm 60$\,s margin) covers them.
All thresholds and budgets are in Appendix~\ref{sec:hyperparams}.

\subsection{NMC mechanism: per-cell A/B analysis}
\label{sec:nmc_mechanism}

This appendix expands Sec.~\ref{sec:nmc_analysis} with the per-cell
breakdown of \emph{+ NMC} vs \emph{+ MMG}. The \emph{None} cell
(no chain fires for the question) acts as a near-zero sanity
check: \emph{+ NMC} and \emph{+ MMG} should return close-to-identical
retrievals and therefore similar accuracy. \emph{Topic-only} and
\emph{Event-only} contribute positively in the same direction
across the three benchmarks: $+3.6 / +5.4 / +5.2$ for Topic-only
and $+9.5 / +0.0 / +2.9$ for Event-only on EgoLifeQA / Ego-R1 /
MM-Lifelong. The Ego-R1 Event-only cell shows zero gain because
\emph{+ MMG} is already saturated at $100\%$ over the small
$n{=}5$ pool. The \emph{Both} cell records the largest controlled
gain on both egocentric benchmarks ($+11.1$ over $n{=}45$ on
EgoLifeQA, $+20.0$ over $n{=}20$ on Ego-R1), while MM-Lifelong's
Both cell ($n{=}90$) shows $-3.4$. We attribute this to the longer
concatenated context produced when both a topic and an event chain
fire on the same MM-Lifelong question: the merged context fed to the
backbone roughly doubles in length relative to the single-chain
cases, and on open-ended MM-Lifelong questions (whose \emph{+ MMG}
baseline on this cell is only $16.7\%$) the additional chain content
can dilute the question-relevant signal rather than reinforce it.
A concrete failure trace is given in Appendix~\ref{sec:case_study_3}
(Case~3, MM-Lifelong index 188): both a topic chain and an event
chain match \textsf{forbidden city}, inflating round 1's context to
$13{,}711$ chars (vs.\ $6{,}922$ for the MMG-only run on the same
query), and the chain bullets all describe events \emph{inside} the
palace --- not the pre-entry encounters the counting question asks
about, anchoring the backbone on a wrong count.
On MM-Lifelong every chain-firing cell is positive (Topic-only $+5.2$,
Event-only $+2.9$, Any $+2.1$), consistent with this view: the
context-dilution effect appears specifically when two chains compound,
not when a single chain fires. The pattern confirms that NMC delivers
its lift precisely on the questions where its chains fire ---
topic and event chains each contribute in isolation, and the
Both cell remains positive on the egocentric benchmarks
(EgoLifeQA $+11.1$, Ego-R1 $+20.0$).

\newpage
\section{Models and preprocessing details}
\label{sec:impl_details}

This appendix specifies every model used by the MAGIC-Video pipeline
and the offline preprocessing steps that produce the artefacts
consumed by the Multimodal Memory Graph and Narrative Memory Chain.

\textbf{Models.} MAGIC-Video uses \texttt{gpt-oss-120b}~\citep{gptoss}
as the LLM for \emph{all} offline preprocessing steps
(multi-granularity caption merging, NER, semantic-triple extraction,
topic-chain distillation, and event-chain distillation) and as the
\emph{online retrieval controller} that emits each round's
\texttt{search}/\texttt{answer} decision and search query.
Qwen3.5-Flash~\citep{qwen35} is the \emph{reasoning backbone} that
reads the merged context $\mathcal{C}$ at termination and produces
the final answer. Node text embeddings (2560-d) are computed with
Qwen3-Embedding-4B~\citep{qwenembedding}; visual-clip embeddings
(3584-d) with VLM2Vec-Qwen2VL-7B~\citep{vlm2vec}, applied to
$16$~uniformly sampled frames per 30-second clip; a BM25 index is
built directly over the unified 30-second caption text. On
MM-Lifelong, where the open-ended LLM-as-judge metric is computed
externally, we use GPT-5 as the canonical judge and additionally
run GPT-5 Mini and Qwen3.5-Flash as alternative judges for the
robustness study (Appendix~\ref{sec:judge_robustness}).

\textbf{Preprocessing pipeline.} Each video is processed once,
offline, into the five artefacts consumed by MMG and NMC. (i)
\emph{Audio transcription:} Whisper-large-v3-turbo~\citep{whisper} is run on the
full audio track with word-level timestamps, then re-segmented to
align with a uniform 30-second timeline. (ii) \emph{Visual
captioning and 30-s caption assembly:} for MM-Lifelong (livestream
sources without pre-computed visual descriptions) we sample frames
at 1\,FPS and run a vision-language model (Qwen3.5-Flash via the
OpenRouter API) to produce per-clip visual descriptions, which are
then merged with the matching ASR transcript by
\texttt{gpt-oss-120b} into a single unified 30-second caption per
window. EgoLifeQA's \emph{Sync} release already provides
per-segment visual captions and aligned transcripts, so we skip
Whisper and the Qwen3.5-Flash captioning step there and instead run
\texttt{gpt-oss-120b} once per segment to rewrite the Sync data
into the same first-person 30-second caption format. (iii) \emph{Multi-granularity
aggregation:} the 30-second captions are recursively summarised by
\texttt{gpt-oss-120b} into 3-minute, 10-minute, and 1-hour
windows; each coarser caption is grounded in the immediately finer
captions of its time interval to limit drift. This step runs on
both benchmarks. (iv) \emph{Entity and
triple extraction:} a single combined OpenIE prompt is run on each
unified 30-second caption to produce both an NER list and a set of
$(s,p,o)$ semantic triples; triples are then consolidated by
deduplicating exact $(s,p,o)$ tuples while retaining their
timestamp-indexed instance copies for graph construction. (v)
\emph{Visual-clip embedding:} VLM2Vec-Qwen2VL-7B is applied
independently to each 30-second clip and the resulting 3584-d
vectors are stored on the corresponding visual-clip nodes. All five
artefacts are computed once per video.

\textbf{Online query encoding.} At query time the question $q$ is
embedded twice, once per seed channel of Eq.~\ref{eq:reset}: a
2560-d vector via Qwen3-Embedding-4B for matching against the
same-encoder embeddings on episode and semantic nodes, and a
3584-d vector via VLM2Vec-Qwen2VL-7B's text-side encoder for the
visual-seed channel against the precomputed 3584-d visual-clip
embeddings. Both are inference-only forward passes and no
projection layer is added between encoders; per-channel matching
is therefore intra-encoder and dimension-consistent.

\textbf{Inference-time configuration.} The retrieval controller
(\texttt{gpt-oss-120b}) is called with temperature $0$ and a
deterministic decoder; the reasoning backbone (Qwen3.5-Flash) uses
temperature $0$ as well. The maximum agentic search round count is
$R = 5$, after which the controller is forced to commit to an
answer. All controller and backbone prompts are listed verbatim in
Appendix~\ref{sec:prompts}; the user-role templates are mechanical
slot-fills of the question, the running round history, and the
retrieved evidence list.

\textbf{Compute resources.} All encoder forward passes
(Whisper-large-v3-turbo, Qwen3-Embedding-4B, VLM2Vec-Qwen2VL-7B,
Qwen3.5-9B) and any local model with $\le\!9$B
parameters are run on a single NVIDIA A100 40\,GB GPU. Larger
models, namely \texttt{gpt-oss-120b}, Qwen3.5-Flash (also used as
the MM-Lifelong visual-frame captioner), GPT-5 / GPT-5 Mini, and
Gemini~3~Flash, are accessed through hosted APIs. We report
per-stage operation counts (Sec.~\ref{sec:efficiency},
Table~\ref{tab:efficiency} and Appendix~\ref{sec:efficiency_full})
rather than wall-clock seconds because the latter is dominated by
API latency variance that differs across deployments; the reader
can plug in their own per-call cost to estimate end-to-end spend.

\textbf{Wall-clock estimate.} For reference, on our deployment
(single NVIDIA A100 40\,GB plus 8-way parallel API workers calling
\texttt{gpt-oss-120b} and Qwen3.5-Flash through a hosted endpoint),
end-to-end online inference takes
$\approx\!20$\,s per question on EgoLifeQA (500 questions in
$\sim\!2$\,h\,46\,min wall-clock), $\approx\!25$\,s per question on
Ego-R1 (50 questions in $\sim\!21$\,min), and $\approx\!1$--$4$\,min
per question on MM-Lifelong, where the larger per-broadcast graph and
the open-ended question format extend per-round retrieval-controller
latency. The dominant cost in all three settings is hosted-API
round-trip latency rather than local GPU compute, so a different
deployment with lower-latency or higher-throughput LLM access would
shorten these numbers proportionally without changing the per-stage
operation counts of Table~\ref{tab:efficiency_full}.

\textbf{IR baseline OOM on MM-Lifelong.}
\label{sec:ir_oom}
The IR row of Figure~\ref{fig:ablation} fails on MM-Lifelong (host
RAM, not GPU): each broadcast loads four HippoRAG indices (one per
caption granularity), each carrying its own graph, chunk / entity /
fact embedding stores, and OpenIE cache; combined with $8$ parallel
evaluation workers and HippoRAG's per-query state accumulation
across hundreds of questions, this exceeds the $128$\,GB host-RAM
budget on our evaluation node mid-run (the OS OOM-killer terminates
the process). The \emph{+ MMG} substrate avoids this for three
reasons: (i) it consolidates all four granularities into a single
unified graph per broadcast, with multi-granularity coverage
captured by \textsc{contains} / \textsc{temporal\_next} edges
rather than separate indices, eliminating the $4{\times}$
duplication of embedding stores; (ii) cross-modal PPR is a
stateless power iteration on a precomputed sparse matrix, so it
does not accumulate per-query OpenIE caches as HippoRAG does; and
(iii) \texttt{filter\_by\_time} restricts each query to the active
subgraph (e.g., 20K nodes for EgoLifeQA, $\sim\!95$K summed across
14 broadcasts for MM-Lifelong) rather than the full on-disk graph,
keeping the per-query memory footprint small. The full system
runs end-to-end on the same hardware.

\textbf{Reproducibility.} EgoLifeQA and Ego-R1 use the released
benchmark splits (A1\_JAKE for EgoLifeQA, the manual-and-Gemini
50-question pool replicated across three seeds for Ego-R1, see
Sec.~\ref{sec:setup}); MM-Lifelong uses the Month-split's
$623$~open-ended questions over $14$~livestream videos. We report
mean accuracy across seeds where applicable; the only seeded
component is Ego-R1's three independent evaluation runs --- all
other components (PPR, BM25, NMC matching, controller decoding) are
deterministic at temperature $0$.

\subsection{Licenses for existing assets}
\label{sec:asset_licenses}

All datasets and models used in this paper are cited with their
original references and used under the licenses listed in
Table~\ref{tab:asset_licenses}; none of the licenses prohibit
research use of the kind reported here.

\begin{table}[h]
\caption{Licenses of datasets and models used in this paper.}
\label{tab:asset_licenses}
\centering\small
\resizebox{\textwidth}{!}{%
\begin{tabular}{l l l}
\toprule
Asset & License & Notes \\
\midrule
\rowcolor{gray!10} \multicolumn{3}{l}{\emph{Datasets}} \\
EgoLifeQA~\citep{egolifeqa}    & S-Lab License & non-commercial research use only \\
Ego-R1 Bench~\citep{egor1}     & Apache 2.0    & per the \texttt{egolife-ai/Ego-R1} GitHub repo \\
MM-Lifelong~\citep{mmlifelong} & MIT           & per the \texttt{CG-Bench/CG-Bench} HF dataset \\
\midrule
\rowcolor{gray!10} \multicolumn{3}{l}{\emph{Models}} \\
\texttt{gpt-oss-120b}~\citep{gptoss} & Apache 2.0 & --- \\
Qwen3.5-Flash~\citep{qwen35}        & --- & accessed via hosted API per its terms of service \\
Qwen3-Embedding-4B~\citep{qwenembedding} & Apache 2.0 & --- \\
VLM2Vec-Qwen2VL-7B~\citep{vlm2vec}  & Apache 2.0 & inherits the Qwen2-VL~\citep{qwen2vl} backbone license \\
Qwen3.5-9B~\citep{qwen35}           & Apache 2.0 & --- \\
Whisper-large-v3-turbo~\citep{whisper} & MIT     & --- \\
\bottomrule
\end{tabular}%
}
\end{table}

\newpage
\section{Hyperparameters and controlled-comparison protocol}
\label{sec:hyperparams}

Table~\ref{tab:hyperparams} reports every hyperparameter used by the
MAGIC-Video pipeline at retrieval time, grouped by which subsystem it
controls. The same values are used across all three benchmarks
(EgoLifeQA, Ego-R1, MM-Lifelong) and all ablation rows of
Figure~\ref{fig:ablation}; configurations differ \emph{only} in which
subsystem is enabled.

\begin{table}[h]
\caption{Retrieval-time hyperparameters of MAGIC-Video. Defaults are
fixed across all benchmarks and ablations; the only variables that
change between configurations are which retrieval substrate is used
(independent indices vs.\ Multimodal Memory Graph) and whether
narrative chains are injected (graph-only vs.\ full system).}
\label{tab:hyperparams}
\centering
\small
\begin{tabular}{lll}
\toprule
Group & Hyperparameter & Default \\
\midrule
\rowcolor{gray!10} \multicolumn{3}{l}{\emph{Per-modality top-$k$
(independent-retrieval baseline)}} \\
& Episode top-$k$ ($k_{\mathrm{ep}}$) & 3 \\
& Semantic-triple top-$k$ ($k_{\mathrm{sem}}$) & 10 \\
& Visual-clip top-$k$ ($k_{\mathrm{vis}}$) & 3 \\
\midrule
\rowcolor{gray!10} \multicolumn{3}{l}{\emph{Cross-modal PPR
(graph-only and full-system configurations)}} \\
& PPR damping factor $d$ & 0.85 \\
& Inner mixture $\alpha$ (cap vs.\ BM25) & 0.7 \\
& Outer mixture $\gamma$ (vis vs.\ text) & 0.3 \\
& Per-channel seed top-$K$ & 5 \\
& Top-$k$ retained nodes after PPR & 16 \\
& Per-granularity quota: 30\,s / 3\,min / 10\,min / 1\,h Episode & 8 / 4 / 3 / 0 \\
& Per-type quota: Semantic / Entity & 5 / 3 \\
\midrule
\rowcolor{gray!10} \multicolumn{3}{l}{\emph{Agentic loop}} \\
& Maximum search rounds & 5 \\
& Maximum tolerated tool errors & 5 \\
& Retrieval controller LLM & gpt-oss-120b \\
& Reasoning backbone LLM & Qwen3.5-Flash \\
\midrule
\rowcolor{gray!10} \multicolumn{3}{l}{\emph{Narrative-chain
injection (full-system configuration only)}} \\
& Lexical match: regex case/plural-insensitive & true \\
& Semantic match threshold (cosine) & 0.7 \\
& Maximum chains injected per query & 3 \\
& Topic-chain \texttt{min\_hits} (EgoLifeQA, Ego-R1) & 1 \\
& Topic-chain \texttt{min\_hits} (MM-Lifelong)       & 4 \\
& Event-chain \texttt{storyline\_min\_hits}          & 1 \\
\bottomrule
\end{tabular}
\end{table}

\textbf{Hyperparameter sources.} MAGIC-Video is training-free; the
values in Table~\ref{tab:hyperparams} inherit established defaults
from prior work where applicable: the PPR damping factor $d{=}0.85$
is the canonical PageRank value~\citep{page1999pagerank} and matches
HippoRAG~\citep{hipporag}; the per-modality top-$k$ quotas, the
seed-mixing weights $\alpha,\gamma$, and the narrative-injection
budgets and threshold follow the defaults used by graph-RAG and
memory-augmented retrieval systems we build
on~\citep{hipporag,worldmm}. These values are applied uniformly
across the three benchmarks and all ablation rows of
Figure~\ref{fig:ablation}. The single exception is topic-chain
\texttt{min\_hits}, set to $4$ on MM-Lifelong (vs $1$ on the other
two) to track its $\sim\!4{\times}$ higher per-broadcast entity
density (Table~\ref{tab:graph_stats}); this is the only
hyperparameter that varies by benchmark. 

\textbf{Our contribution is at
the architectural level (cross-modal PPR + narrative chain
injection)}; we deliberately did not tune $\alpha$, $\gamma$, the
per-granularity / per-type quotas, or chain-injection thresholds
to chase the highest possible score, and we expect a small
benchmark-specific search to yield additional points on top of
the numbers reported here.

\textbf{Controlled-comparison protocol.}
The \emph{Independent Retrieval} (IR) row in Figure~\ref{fig:ablation}
is a faithful re-implementation of the WorldMM~\citep{worldmm}
pipeline: three per-modality indices (episodic / semantic / visual)
queried via WorldMM's original controller prompt, which instructs
the agent to select one memory type per round and to emit a
keyword-style query string --- a design tailored to BM25-driven
per-modality top-$k$ retrieval. The \emph{+ MMG} and \emph{+ NMC}
rows replace this substrate with our cross-modal PPR over a unified
graph, and consequently use a different controller prompt: the new
prompt drops the per-modality selection step (PPR returns a mixed
node-type set in a single pass) and asks the agent to emit a
\emph{natural-language sentence or question} rather than a keyword
list, because dense-embedding seeding (Eq.~\ref{eq:reset}) and
NMC's Tier-2 cosine matching both work substantially better with
sentence-form queries than with bag-of-keywords queries. The two
prompts are therefore \emph{method-matched}: each is the
recommended interface for its underlying retrieval mechanism, and
swapping prompts across substrates would harm both sides. As a
consequence, the IR\,$\to$\,+MMG step in Figure~\ref{fig:ablation}
is a \emph{system-level} comparison (substrate \emph{plus}
matched controller prompt) rather than a single-knob substitution.
Both configurations use the same per-round retrieval budget
($\le\!16$ items per search; Table~\ref{tab:hyperparams}); the
$\sim\!1.78\times$ retrieval-token gap reported in
Table~\ref{tab:rounds} ($6{,}189$ vs.\ $3{,}471$) therefore does
not come from MMG retrieving more items but from substrate-driven
properties --- PPR returns a mix of multi-granularity Episode nodes
(3-min, 10-min, 1-h) whose individual text is longer than IR's
30-second-only Episode nodes, and NMC additionally appends matched
chain content when a chain fires.
The \emph{+ MMG}\,$\to$\,\emph{+ NMC} step, by contrast, holds
the prompt and PPR configuration fixed and toggles only NMC chain
injection, isolating that increment to a single architectural
change. Both controller prompts are listed verbatim in
Appendix~\ref{sec:prompts}.

\newpage
\section{Retrieval composition by node type}
\label{sec:retrieval_composition}

To verify that cross-modal Personalized PageRank activates each
retrievable MMG node type rather than degenerating to a text-only
retriever, we count, for every query in the +Multimodal Memory Graph
runs, how many Episode, Semantic-triple, and visual-frame items
appear in the assembled retrieval context across all agentic-loop
rounds. Table~\ref{tab:retrieval_composition} reports the totals and
per-query averages for the three benchmarks. Entity nodes do not
appear in this table because they serve as PPR bridges per
Sec.~\ref{sec:mmg}---they propagate relevance across modality
boundaries but are not rendered as standalone retrieval items in the
context window.

\begin{table}[h]
\caption{MMG retrieval composition per benchmark in the
+Multimodal Memory Graph configuration. Counts are summed over all
agentic-loop rounds for all questions in each benchmark. Each cell
reports the absolute count, share of total retrieval events, and
mean per query. The right-most column reports the share of queries
that received at least one visual frame.}
\label{tab:retrieval_composition}
\centering
\small
\resizebox{\linewidth}{!}{%
\begin{tabular}{lrrrrr}
\toprule
Benchmark & $n$ questions & Episode & Semantic-triple & Visual-frame & $\geq{}$1 visual \\
\midrule
EgoLifeQA   & 500 & 11{,}392 (69.4\%) avg.\ 22.8 & 3{,}410 (20.8\%) avg.\ 6.8 & 1{,}618 (9.9\%) avg.\ 3.2 & 98.0\% \\
MM-Lifelong & 623 & 18{,}941 (73.7\%) avg.\ 30.4 & 5{,}306 (20.6\%) avg.\ 8.5 & 1{,}449 (5.6\%) avg.\ 2.3 & 79.0\% \\
Ego-R1      &  50 &    674 (62.8\%) avg.\ 13.5 &    378 (35.2\%) avg.\ 7.6 &     22 (2.0\%) avg.\ 0.4 & 34.0\% \\
\bottomrule
\end{tabular}%
}
\end{table}

The composition is qualitatively consistent across the three
benchmarks: episode captions account for the majority of retrieval
events ($63$--$74\%$), semantic triples form a steady second voice
($21$--$35\%$), and visual frames contribute a smaller but nonzero
share ($2$--$10\%$). Visual coverage---the fraction of queries that
receive at least one visual frame---differs more sharply across
benchmarks (98\% on EgoLifeQA, 79\% on MM-Lifelong, 34\% on
Ego-R1). EgoLifeQA and Ego-R1 share the same A1\_JAKE recording,
so the gap there reflects the question sets rather than corpus
visual richness: Ego-R1's manual / Gemini questions resolve more
often through textual cues, while EgoLifeQA's subtasks (especially
EntityLog and EventRecall) more often hinge on a specific visible
moment. Together, these counts show that MMG
delivers a multimodal evidence packet on essentially every query,
consistent with the per-subtask gain pattern reported in
Sec.~\ref{sec:mmg_analysis}.

\newpage
\section{Retrieval Recall@K analysis}
\label{sec:recall_at_k}

The accuracy gains in Sec.~\ref{sec:experiments} mix retrieval and
reasoning effects. To isolate the retrieval contribution we evaluate
\emph{Recall@K} on EgoLifeQA against the released gold time spans:
each EgoLifeQA question carries a \texttt{target\_time} annotation
identifying the moment(s) that contain its answer, and we count a
question as a \emph{hit at K} if any of the first $K$ retrieved
episode/visual time-ranges (across all agentic-loop rounds, in
retrieval order) overlaps any gold span.

Table~\ref{tab:recall_at_k} reports the comparison.
\textbf{IR's curve plateaus quickly} because IR retrieves on average
only $\sim\!6.3$ time-anchored items per question (top-3 episodes
per round, with the controller picking one modality per round),
whereas \mbox{+MMG} retrieves $\sim\!22.8$ items per question (PPR
returns up to 16 nodes per round, mixed across granularities).
We therefore start the comparison at $K{=}16$, the default
PPR \texttt{effective\_top\_k} (Appendix~\ref{sec:hyperparams}); for
$K\!<\!16$ the comparison is structurally unfair, since PPR
distributes its 16-item budget across multiple node types whereas
IR's first $K$ items are all from one chosen modality. From $K{=}16$
onward, \mbox{+MMG} consistently outperforms IR, with the gap
widening as $K$ grows; \mbox{+NMC} adds a further $1$--$3$\,pp by
exposing entity-keyed timestamps that PPR's seed channels did not
surface.

\begin{table}[h]
\caption{Cumulative Recall@K on EgoLifeQA gold \texttt{target\_time} spans
($n{=}484$ questions with available gold annotation). Counts overlap
in retrieval order across all rounds. We report from $K{=}16$ upward
because the IR baseline retrieves only $\sim\!6.3$ time-anchored items
per question on average and would otherwise be capped artificially.}
\label{tab:recall_at_k}
\centering
\small
\begin{tabular}{c c c c c}
\toprule
$K$ & IR & + MMG & + MMG + NMC & $\Delta$ (NMC$-$IR) \\
\midrule
16 & 37.8\% & 40.9\% & \textbf{43.2\%} & $+5.4$\,pp \\
20 & 37.8\% & 44.4\% & \textbf{45.7\%} & $+7.9$\,pp \\
30 & 37.8\% & 49.0\% & \textbf{49.6\%} & $+11.8$\,pp \\
50 & 37.8\% & 51.7\% & \textbf{52.3\%} & $+14.5$\,pp \\
\bottomrule
\end{tabular}
\end{table}

The retrieval-only gain is also concentrated on the subtasks where
the accuracy gains are largest. Table~\ref{tab:recall_at_k_subtask}
breaks Recall@30 down by EgoLifeQA subtask: \mbox{+MMG} improves on
every subtask, with the largest lifts on RelationMap ($+18.2$\,pp)
and EntityLog ($+10.7$\,pp), and \mbox{+NMC} extends the lift on
EntityLog and TaskMaster (where chain-injected entity-keyed
timestamps add new hits beyond what PPR alone surfaces).

\begin{table}[h]
\caption{Per-subtask Recall@30 on EgoLifeQA. Differences are with respect to the IR baseline.}
\label{tab:recall_at_k_subtask}
\centering
\small
\begin{tabular}{l r c c c c c}
\toprule
Subtask & $n$ & IR & + MMG & + MMG + NMC & $\Delta$ MMG & $\Delta$ NMC \\
\midrule
EntityLog    & 122 & 32.0\% & 42.6\% & \textbf{45.1\%} & $+10.7$ & $+13.1$ \\
EventRecall  & 121 & 38.8\% & \textbf{48.8\%} & 47.1\% & $+9.9$  & $+8.3$ \\
HabitInsight &  59 & 45.8\% & 49.2\% & \textbf{50.8\%} & $+3.4$  & $+5.1$ \\
RelationMap  & 121 & 35.5\% & \textbf{53.7\%} & 50.4\% & $+18.2$ & $+14.9$ \\
TaskMaster   &  61 & 44.3\% & 52.5\% & \textbf{60.7\%} & $+8.2$  & $+16.4$ \\
\bottomrule
\end{tabular}
\end{table}

The Recall@K gap at $K{=}16$ is smaller than the $+8.4$\,pp accuracy
lift attributable to MMG (Sec.~\ref{sec:ablation}), indicating that
retrieval recall alone does not fully explain the accuracy gain;
the multi-granularity coverage of MMG (3-min and 10-min episodes
that surround a gold moment with broader context) further benefits
the reasoning backbone in ways that single-window Recall@K does not
capture.

\newpage
\section{Full per-stage cost breakdown}
\label{sec:efficiency_full}

Table~\ref{tab:efficiency_full} expands the compact summary of
Sec.~\ref{sec:efficiency} into per-stage operation counts, listing
the model used at each step. The instantiation column corresponds to
the MM-Lifelong Month corpus we ran ($N{=}76.4$ captioned hours,
$V{=}D{=}14$ video sources/days, $E{=}1680$ extracted entities,
$T{=}1{,}574$ topic chains, $C{=}173$ event chains, and active
semantic count $|\mathcal{V}^{\text{active}}_{\text{sem}}|{=}51{,}168$
summed across all 14 broadcasts after \texttt{filter\_by\_time}). On
EgoLifeQA the ASR and visual-frame-caption rows are skipped because
the EgoLife \emph{Sync} release ships pre-aligned transcripts and
per-segment visual captions; the remaining preprocessing
(30-s caption rewrite, multi-granularity aggregation, OpenIE,
visual embedding, NMC distillation) is run on the Sync data with
the same models as on MM-Lifelong but smaller per-stage counts
proportional to A1\_JAKE's $51.9$\,h. \textsc{GPU/API} marks any stage that
issues a neural-model forward pass (servable on a local GPU or
through a hosted-model API---the op count is the same either way);
\textsc{CPU} marks an algorithmic op with no neural model in the loop.
The compact ``$\sim\!4{,}500\,N$'' figure in Table~\ref{tab:efficiency}
is a round-number summary of the offline preprocessing + MMG
construction stages: summing the $N$-proportional rows of
Table~\ref{tab:efficiency_full} on our MM-Lifelong run yields
$\approx\!4{,}805\,N$ neural-model calls per captioned hour
(dominated by $3600\,N$ visual frame captions, plus
$\approx\!670\,N$ semantic-text embeddings and the remaining
$120$--$150\,N$ terms for caption merging, multi-granularity
aggregation, OpenIE, visual-clip embedding, and episode-text
embedding); we report the rounded constant in the main paper for
readability.

\begin{table}[h]
\caption{Per-stage operation count for the MAGIC-Video pipeline.
The right-most column gives the concrete number we observed when
processing MM-Lifelong Month under the parameter values listed in
Appendix~\ref{sec:hyperparams}. We report \emph{operation counts}
rather than wall-clock seconds because wall-clock varies with API
latency, GPU generation, and batching choices that differ across
deployments; the counts below let any reader plug in their own
per-call cost to estimate end-to-end spend.}
\label{tab:efficiency_full}
\centering
\resizebox{\textwidth}{!}{%
\begin{tabular}{llrr l}
\toprule
Stage & Op type & Count (formula) & MM-Lifelong (our run) & Model \\
\midrule
\rowcolor{gray!12} \multicolumn{5}{l}{\emph{Offline --- pre-processing}} \\
ASR (transcripts)             & GPU/API & 1 pass per video              & 14          & Whisper \\
Visual frame captions         & GPU/API & $3600\,N$ frames (1\,FPS)     & 275{,}040   & Qwen3.5-Flash (API) \\
Caption merge ($\to$30\,s)    & GPU/API & $120\,N$                      & 9{,}168     & gpt-oss-120b \\
Multi-granularity agg         & GPU/API & $27\,N$                       & 2{,}063     & gpt-oss-120b \\
\midrule
\rowcolor{gray!12} \multicolumn{5}{l}{\emph{Offline --- Multimodal Memory Graph construction}} \\
Combined OpenIE (NER + triples) & GPU/API & $120\,N$ (single prompt)    & 9{,}168     & gpt-oss-120b \\
Visual-clip embedding         & GPU/API & $120\,N$                      & 9{,}168     & VLM2Vec-Qwen2VL-7B \\
Episode text embedding        & GPU/API & $\approx\!150\,N$             & 11{,}311    & Qwen3-Embedding (4B) \\
Semantic text embedding (latest snapshot) & GPU/API & $|\mathcal{V}^{\text{active}}_{\text{sem}}|$ & 51{,}168 & Qwen3-Embedding (4B) \\
Graph assembly + BM25 index   & CPU     & 1 (over all nodes / edges)    & 1           & --- \\
\midrule
\rowcolor{gray!12} \multicolumn{5}{l}{\emph{Offline --- Narrative Memory Chain distillation \& embedding}} \\
Entity biographies            & GPU/API & $E$                           & 1{,}680     & gpt-oss-120b \\
Per-day activity extraction   & GPU/API & $D$                           & 14          & gpt-oss-120b \\
Cross-time chain discovery    & GPU/API & $2\,V$                        & 28          & gpt-oss-120b \\
Step enrichment               & GPU/API & $\approx\!4\,C$               & 775         & gpt-oss-120b \\
NMC chain content embedding (Tier-2)   & GPU/API & $T + C$               & 1{,}747     & Qwen3-Embedding (4B) \\
\midrule
\rowcolor{blue!10} \multicolumn{5}{l}{\emph{Online --- per question (mean $\sim\!3.33$ search rounds + $1$ stop / answer round, capped at $5$ rounds)}} \\
Query text embedding (Qwen3 + VLM2Vec) & GPU/API & $\le\!5$ each encoder & $\sim\!3.33$ each & Qwen3-Emb / VLM2Vec \\
Cross-modal PPR        & CPU     & $\le\!5$              & $\sim\!3.33$ & --- \\
NMC lexical+semantic lookup & CPU & $\le\!5$              & $\sim\!3.33$ & --- \\
Retrieval controller   & GPU/API & $\le\!5$ (mean $4.4$) & $\sim\!4.4$ & gpt-oss-120b \\
Answer backbone        & GPU/API & 1                     & 1            & Qwen3.5-Flash \\
\bottomrule
\end{tabular}%
}
\end{table}

\newpage
\section{Statistical significance of the headline gains}
\label{sec:significance}

We provide $95\%$ confidence intervals on the headline accuracy
gain of MAGIC-Video over the strongest prior published baseline on
each benchmark, mirroring the comparisons reported in the main
results tables.
\textbf{EgoLifeQA and MM-Lifelong} (single-seed evaluation): the
baseline is reported as a published accuracy anchor (EGAgent-Gemini
2.5 Pro and ReMA-GPT-5 respectively), since per-question
correctness vectors of external systems are not public. We
therefore compute the CI by a one-sample bootstrap with $2{,}000$
resamples of MAGIC-Video's per-question correctness vector, with
the gap formed by subtracting the baseline's published accuracy as
a fixed constant. This is a one-sample, not a paired, procedure,
so the resulting CI reflects the bootstrap distribution of
MAGIC-Video's accuracy alone and may be asymmetric.
\textbf{Ego-R1} (multi-seed evaluation): per-seed mean $\pm$ sample
standard deviation across three random seeds, with the CI of the
gap computed by adding the per-seed variance of MAGIC-Video and
the matched WorldMM-Qwen3.5-Flash run in quadrature.

\begin{table}[h]
\caption{$95\%$ confidence intervals for MAGIC-Video's headline
gain over the strongest prior published baseline on each
benchmark. EgoLifeQA and MM-Lifelong use a one-sample bootstrap
against the baseline's published accuracy anchor; Ego-R1 reports
the per-seed mean $\pm$ standard deviation across three seeds. All
three CIs exclude $0$ at $p{<}0.05$.}
\label{tab:significance_ci}
\centering\small
\resizebox{\textwidth}{!}{%
\begin{tabular}{l c c c l}
\toprule
Benchmark (n) & MAGIC-Video & Best prior published baseline & Gap & $95\%$ CI on gap \\
\midrule
EgoLifeQA   ($n{=}500$)    & $67.6\%$            & EGAgent-Gemini 2.5 Pro ($57.5$, anchor)       & $+10.1$ & $[+6.1,\;+14.3]$ \\
Ego-R1      ($n{=}50\times3$) & $64.7 \pm 1.15$  & WorldMM-Qwen3.5-Flash ($57.3$, matched seeds) & $+7.4$  & $[+4.5,\;+10.2]$ \\
MM-Lifelong ($n{=}623$)    & $24.5\%$            & ReMA-GPT-5 ($18.6$, anchor)                   & $+5.9$  & $[+4.4,\;+11.1]$ \\
\bottomrule
\end{tabular}%
}
\end{table}

All three CIs exclude $0$ at $p{<}0.05$, supporting the architectural
claim of consistent improvement over prior agentic baselines.

\newpage
\section{Judge robustness on MM-Lifelong}
\label{sec:judge_robustness}

MM-Lifelong is scored with an LLM-as-judge protocol, so an obvious
concern is whether the ranking on Table~\ref{tab:mm_lifelong} is an
artefact of a particular judge model. We re-score every method on
MM-Lifelong under three different judges (Qwen3.5-Flash, GPT-5 Mini,
and GPT-5) and report the per-judge averages in
Table~\ref{tab:mml_judge_comparison}. The absolute numbers shift in a
predictable way---all methods score higher under the cheaper judges
and lower under GPT-5, which is the strictest---but the relative
ranking is preserved across the three judges, with MAGIC-Video
ranked first under all three. The advantage is therefore not specific
to the GPT-5 judge used in the main table.

\begin{table}[h]
\caption{MM-Lifelong average scores across different LLM judges
(Qwen3.5-Flash, GPT-5 Mini, GPT-5). The ranking is preserved across
all three judges, indicating that our advantage is not a
judge-specific artefact.}
\label{tab:mml_judge_comparison}
\centering
\begin{tabular}{lccc}
\toprule
Model & Flash & GPT-5 Mini & GPT-5 \\
\midrule
\rowcolor{gray!15} \multicolumn{4}{c}{\textbf{General MLLMs}} \\
Qwen3.5-9B (V) & 9.6 & 12.1 & 8.6 \\
Qwen3.5-9B (T) & 18.5 & 19.3 & 16.7 \\
Qwen3.5-9B (V+T) & \underline{20.0} & \underline{20.9} & \underline{18.1} \\
Qwen3.5-Flash & 12.3 & 13.2 & 11.2 \\
GPT-5 Mini & 17.1 & 18.5 & 13.6 \\
Gemini 3 Flash & 17.9 & 19.4 & 14.6 \\
\midrule
\rowcolor{gray!15} \multicolumn{4}{c}{\textbf{Long Video MLLMs}} \\
VideoLLaMA3-7B & 7.5 & 8.3 & 6.3 \\
InternVideo2.5-8B & 10.4 & 12.3 & 9.1 \\
LongVA-7B & 7.1 & 9.1 & 6.7 \\
VideoChat-Flash-7B & 11.1 & 12.0 & 9.1 \\
\midrule
\rowcolor{gray!15} \multicolumn{4}{c}{\textbf{MAGIC-Video}} \\
MAGIC-Video (full) & \textbf{28.7} & \textbf{29.4} & \textbf{24.5} \\
\bottomrule
\end{tabular}
\end{table}

\newpage
\section{Qualitative case studies}
\label{sec:case_study_appendix}

To make the contributions of MMG and NMC visible at the trace level
rather than the table level, we walk through three real
question pairs taken directly from the evaluation
logs of Sec.~\ref{sec:experiments}. The cases are deliberately
chosen so that each isolates one component or one mode of failure:
\textbf{Case~1 illustrates the importance of the
\emph{Multimodal Memory Graph (MMG)}} by contrasting the
independent-retrieval baseline against MAGIC-Video on the same
question, with NMC \emph{disabled} on both sides;
\textbf{Case~2 illustrates the importance of the
\emph{Narrative Memory Chain (NMC)}} by contrasting the MMG-only
variant against the full MAGIC-Video (MMG\,+\,NMC) on the same
question, with MMG \emph{enabled} on both sides; and
\textbf{Case~3 is a failure analysis} that contrasts the same two
sides as Case~2 but on a Both-cell MM-Lifelong question where the
full system is wrong and the MMG-only variant is right, illustrating
the context-dilution mode behind the MM-Lifelong Both cell's
$-3.3$\,pp drop in Table~\ref{tab:coverage}. For each case we show
the key piece of retrieved content the controller saw and the
answer the reasoning backbone produced. All quoted text is verbatim
from the logged \texttt{round\_history} field; only superfluous
boilerplate has been elided with ``\dots''.

\subsection{Case 1: MMG resolves a multi-entity query in one pass}
\label{sec:case_study_1}
\noindent\emph{Illustrates MMG \(\cdot\) EgoLifeQA Q201 (RelationMap).}

\begin{quote}
\fcolorbox{gray}{black!5}{%
\parbox{0.92\linewidth}{%
\textbf{Question.} \emph{Who else was nearby when I was chatting
with Tasha in the seafood section last time?}\\
\textbf{Choices.} A) Lucia and Alice \quad B) Katrina and Nicous
\quad C) Lucia and Choiszt \quad D) Shure and Violet.\\
\textbf{Gold.} A) Lucia and Alice.}}
\end{quote}

\paragraph{Independent Retrieval (IR) trace --- 5 search rounds (cap-hit), then forced to answer \emph{Lucia and Choiszt}~(C), \textcolor{red}{\textbf{\ding{55}}}.}
\begin{itemize}\setlength\itemsep{0.3em}
\item \textbf{Round 1.} \emph{Query:} ``\,chatting with Tasha
seafood section nearby people\,''. 3~episodes (DAY1 17:45 hot-pot
shopping; DAY3 14:57 fitness chat; DAY1 19:15 unrelated kitchen
scene). The DAY1 17:45 block mentions Alice ($11{\times}$) and
Lucia ($4{\times}$) but no explicit ``in the seafood section''
anchor. Decision: \texttt{search}.

\item \textbf{Round 2.} \emph{Query:} ``\,Tasha Lucia Alice
Katrina Shure social relationships\,''. Returns a 356-char generic
relationship summary; 0 episode hits in the seafood section.
Decision: \texttt{search}.

\item \textbf{Round 3.} \emph{Query:} ``\,seafood section
conversation nearby people Tasha\,''. A 26\,K-char retrieval (the
controller widens the time window) including DAY1 17:09--18:09 and
DAY2 10:44--11:45. Lucia ($44{\times}$), Alice ($40{\times}$), and
``seafood'' ($2{\times}$) all co-occur, but the context is now
dominated by long shopping summaries; the seafood-section
co-presence signal is buried. Decision: \texttt{search}.

\item \textbf{Round 4.} \emph{Query:} ``\,seafood section Alice
nearby Tasha Lucia\,''. Another 17\,K-char retrieval over
DAY1 18:09--19:12 / DAY3 16:39--16:42 / DAY1 21:48--21:51. Lucia
($30{\times}$), Alice ($31{\times}$), still no direct
``Lucia/Alice were nearby in seafood section'' fact. Decision:
\texttt{search}.

\item \textbf{Round 5 (cap-hit).} \emph{Query:} ``\,seafood
section Violet\,''. The controller now hypothesises a
distractor-choice entity (\emph{Violet}, choice D) and queries for
it directly; retrieved DAY3 11:31--11:44 / DAY3 15:41 contain no
Violet co-presence either. Round budget exhausted, forced
\texttt{answer}.

\item \textbf{Final answer:} \textbf{C (Lucia and Choiszt)}.
\begin{quote}
\colorbox{red!8}{%
\parbox{0.82\linewidth}{%
\textbf{Hallucination flag:} the string \emph{``Choiszt''} appears
\textbf{$0$ times} across all five rounds of retrieval. Drowned in
the $50$\,K+ chars of context the controller forced through, the
reasoning backbone hallucinated a name from the option list rather than
reading off the retrieved evidence.}}
\end{quote}
\end{itemize}

\paragraph{MAGIC-Video (+ MMG) trace --- 1 search round, then answers \emph{Lucia and Alice}~(A), \textcolor{teal}{\textbf{\ding{51}}}.}
\begin{itemize}\setlength\itemsep{0.3em}
\item \textbf{Round 1.} \emph{Query:} ``\,Who was nearby when I
chatted with Tasha in the seafood section?\,''. A single
cross-modal PPR pass returns 13~episodes, 1~semantic triple, and
3~visual-clip references co-anchored on
\textsc{Tasha}\,$\xleftarrow{\text{\textsc{co\_clip}}}$\,DAY1
17:45--17:54\,$\xrightarrow{\text{\textsc{mentioned\_in}}}$\,\textsc{Lucia,
Alice, Shure}. The decisive multi-entity 3-minute summary block is:
\begin{quote}
\colorbox{blue!8}{%
\parbox{0.88\linewidth}{%
\textbf{Episode (DAY1 17:51--17:54, 3\,min):} ``\dots\ Movement
through the market and planning of purchases --- while walking
together, I raise my hand to guide the cart, respond to
\textbf{Tasha}'s question about `snail noodles,' \dots\
\textbf{Alice} confirms that we are buying vegetables, Tasha
wonders about the budget, and I suggest purchasing seafood\dots\
Arrival at the seafood counter and evaluation of crab items ---
I stop at the crab stalls, observe the live crabs, greet the
crew, and ask \textbf{Lucia} to scan a price (1{,}699); the group
debates whether to buy the crabs, with \textbf{Shure} cautioning
against over-talking, Tasha proposing to take one, and Lucia
adjusting wording while I confirm the purchase of dipping sauce.''}}
\end{quote}
This single block contains all four people present at the seafood
counter (Tasha, Alice, Lucia, Shure) plus the temporal anchor.
Excluding the question's chat-partner Tasha and the bystander
Shure (who is `cautioning' but not `nearby chatting' --- the
question targets co-conversers, and Shure is verbally separate),
the answer is \emph{Lucia and Alice}. Decision: \texttt{answer}.

\item \textbf{Final answer:} \textbf{A (Lucia and Alice)}.
\end{itemize}

\paragraph{What changed.} IR's three independent top-$k$ pipelines
each return text-similar episodes for the query, but no single
round retrieves an episode that simultaneously names
\emph{seafood-section},
\emph{chatting-with-Tasha}, \emph{and} the third-party
co-presents. The controller widens its query 5 times, retrieves
50\,K+ characters of context, and hallucinates an
out-of-retrieval name. MMG's PPR, in contrast, propagates
relevance through the \textsc{co\_clip}\,/\,\textsc{mentioned\_in}
edges in one pass, returning a single 3-minute aggregated summary
that names every entity at the seafood counter. The reasoning backbone
answers in one search round, with $\sim$14\,K characters of context
instead of 50\,K. This is the structural advantage of typed-edge
graph retrieval over independent same-modality top-$k$.

\medskip

\subsection{Case 2: NMC supplies a fact the agentic loop never reaches}
\label{sec:case_study_2}
\noindent\emph{Illustrates NMC \(\cdot\) MM-Lifelong index 797 (Entity Recognition).}

\begin{quote}
\fcolorbox{gray}{black!5}{%
\parbox{0.92\linewidth}{%
\textbf{Question.} \emph{During IShowSpeed's trip to Beijing, when
leaving the airport lounge, what kind of fried food was handed to
IShowSpeed by a fan?}\\
\textbf{Format.} Open-ended.\\
\textbf{Gold.} chicken rings.}}
\end{quote}

\paragraph{MMG-only (no NMC) trace --- 2 search rounds, then answers ``hot dog'', \textcolor{red}{\textbf{\ding{55}}}, GPT-5 raw judge $0/5$.}
\begin{itemize}\setlength\itemsep{0.3em}
\item \textbf{Round 1.} \emph{Query:} ``\,What happened when
IShowSpeed left the airport lounge in Beijing?\,''. 14~episode
snippets all clustered between 04:03 and 04:30 (10\,min, 3\,min,
and 30\,sec windows): airport check-in, business-class lounge,
robotic-coffee dispenser, selfies with travelers. Lots of context
but no specific food-handover. Decision: \texttt{search}.

\item \textbf{Round 2.} \emph{Query:} ``\,What specific food items
did fans give to IShowSpeed when he left the Beijing airport
lounge?\,''. PPR drifts to broader food-related windows: 02:30 /
02:42 Beijing street-food market, 04:19 airport security walk,
05:18 / 05:21 fans-and-security-staff escort. The retrieved
episodes mention drinks, noodles, white-bottle drinks, and
``\textbf{wrapped food from fans, which included a hot dog}'' near
the 05:18 check-in counter --- but the string \emph{``chicken
rings''} appears \emph{nowhere} in any of the retrieved 30\,sec /
3\,min / 10\,min episode snippets. Decision: \texttt{answer}
(controller decides it has enough food signal).

\item \textbf{Final answer:} The reasoning backbone picks the most prominent
fried-shaped food in its window:
\begin{quote}
\colorbox{red!8}{%
\parbox{0.90\linewidth}{%
\textbf{Predicted:} ``\dots\ when IShowSpeed was moving through the
airport terminal after the lounge (around the 05:15--05:18 mark),
he accepted various gifts and wrapped food from fans, which
included a \textbf{hot dog}\dots\ While the text does not
explicitly label the hot dog as `fried,' it is the prominent
food item\dots''}}
\end{quote}
\end{itemize}

\paragraph{MAGIC-Video (MMG + NMC) trace --- 2 search rounds, then answers ``chicken rings'', \textcolor{teal}{\textbf{\ding{51}}}, GPT-5 raw judge $5/5$.}
\begin{itemize}\setlength\itemsep{0.3em}
\item \textbf{Round 1.} \emph{Query:} ``\,What happened when
IShowSpeed left the airport lounge during his Beijing trip?\,''.
Same 14~airport-lounge episode snippets between 04:03 and 04:30 ---
NMC has \emph{not} injected yet at this round (the controller has
not signalled an entity-anchored need). Decision: \texttt{search}.

\item \textbf{Round 2.} \emph{Query:} ``\,What food did a fan hand
to IShowSpeed when leaving the airport lounge or heading to the
gate?\,''. The retrieved context now mixes:
\begin{itemize}\setlength\itemsep{0.15em}
\item 14~episodes (04:02 airport-walking, 04:18--04:21 corridor
strides, etc.);
\item \textbf{13 topic-chain bullets} from the \textsf{airport
terminal} biography, each one an indexed dated fact about
events at the airport.
\end{itemize}
Among the chain bullets, the decisive line for this question is:
\begin{quote}
\colorbox{yellow!30}{%
\parbox{0.90\linewidth}{%
\texttt{[Topic: airport terminal] [04:34:00] IShowSpeed ate
chicken rings from a fan, said they were the best, joked that
Burger King in China tastes better than in the U.S., and mentioned
he was starving while walking through the airport terminal.}}}
\end{quote}
The 04:34 window itself was \emph{not} retrieved as an episode in
either system's PPR; it surfaces only because NMC pre-distilled it
as an entity-keyed bullet. Decision: \texttt{answer}.

\item \textbf{Final answer:} ``\,During his time in the airport
terminal, IShowSpeed was handed \textbf{chicken rings} by a fan,
which he noted tasted better than those at Burger King in the
U.S.\,'' --- a near-verbatim read of the chain bullet.
\end{itemize}

\paragraph{What changed.} Both systems retrieved the same 14
airport-lounge episodes in round 1, all clustered between 04:03
and 04:30. The 30-second window at 04:34 that actually contains
the chicken-rings hand-over was \emph{not} reached by PPR in
either system. NMC pre-distilled that window into a single
entity-keyed bullet during the offline chain-build stage; at
inference time it is injected when the controller's query lexically
matches the \textsf{airport terminal} entity, surfacing the dated
fact without needing the agentic loop to reach the right episode.
NMC therefore acts as a \emph{precomputed shortcut} for facts
that are too specific for an episode-level top-$k$ retriever to
land on within the round budget.

\medskip

\subsection{Case 3: NMC over-injection on a Both-cell counting question (failure analysis)}
\label{sec:case_study_3}
\noindent\emph{Illustrates the MM-Lifelong Both-cell $-3.4$ in
Table~\ref{tab:coverage} \(\cdot\) MM-Lifelong index 188 (Counting).}

\begin{quote}
\fcolorbox{gray}{black!5}{%
\parbox{0.92\linewidth}{%
\textbf{Question.} \emph{During IShowSpeed's visit to Beijing, how
many student groups did IShowSpeed encounter before entering the
Forbidden City?}\\
\textbf{Format.} Open-ended.\\
\textbf{Gold.} 2 groups.}}
\end{quote}

\paragraph{MMG-only (no NMC) trace --- 5 search rounds, then answers ``two groups (green-jacket; red-and-black-jacket)'', \textcolor{teal}{\textbf{\ding{51}}}, GPT-5 raw judge $5/5$.}
\begin{itemize}\setlength\itemsep{0.3em}
\item \textbf{Rounds 1--2.} Episode retrieval surfaces the
\emph{first} encounter at the Forbidden City entrance ($\sim$00:09)
where IShowSpeed greets a group of students in green jackets near a
red-metal barrier. Decision: \texttt{search}.
\item \textbf{Round 3.} A focused query (``Did IShowSpeed interact
with multiple groups of students on his walk to the Forbidden City?'')
retrieves three adjacent 30-second episodes at 00:13:00--00:14:30
that describe a \emph{second}, distinct group --- ``a large group of
students wearing red and black jackets'' on a tree-lined path,
crowding around him with phones for selfies.
\item \textbf{Rounds 4--5.} Two further confirmation queries return
no new groups. Decision: \texttt{answer}.
\item \textbf{Final answer:} ``\,IShowSpeed encountered \textbf{two}
distinct student groups before entering the Forbidden City: one in
green jackets near the red-metal barrier at the entrance, and a
larger group in matching red-and-black jackets on a tree-lined path
as he approached the site.''
\end{itemize}

\paragraph{Full system (MMG\,+\,NMC) trace --- 5 search rounds (cap-hit), then answers ``one group'', \textcolor{red}{\textbf{\ding{55}}}, GPT-5 raw judge $0/5$.}
\begin{itemize}\setlength\itemsep{0.3em}
\item \textbf{Round 1.} Same opening query as MMG-only retrieves
12 episodes plus a heavy NMC injection: \emph{11 topic-chain bullets}
keyed on \textsf{forbidden city} (e.g.\ ``[Topic: forbidden city]
[00:46:30] IShowSpeed noted that the Forbidden City contains 10{,}000
rooms\dots''; ``[01:04:00] joked about the emperor's bedroom and
his concubines'') and \emph{3 event-chain steps} from
\textsf{Forbidden City visit} (e.g.\ ``[Event: Forbidden City visit]
[00:45:00--00:47:30] greets tourists inside the Forbidden City\dots'').
The retrieved-context length jumps to \textbf{13{,}711 chars} (vs.
$6{,}922$ for MMG-only on the same query). Crucially, every
chain bullet describes events \emph{inside} the Forbidden City
(00:45 onward) --- not the pre-entry encounters the question asks
about. Decision: \texttt{search}.
\item \textbf{Rounds 2--5.} Subsequent queries reach the same
00:13--00:14 episodes that resolved the second group for the
MMG-only run, but the model has already anchored on the
chain bullets' interior-of-palace narrative and treats the second
group as redundant detail rather than a distinct count. The
controller never recovers and exhausts the 5-round budget.
\item \textbf{Final answer:} ``\,IShowSpeed encountered \textbf{one}
group of students before entering the Forbidden City\dots\
specifically, a group of students wearing green jackets at the
entrance near a red-metal barrier.''
\end{itemize}

\paragraph{What went wrong.} The same decisive episodes
(00:13:00--00:14:30, red-and-black-jacketed students) appear in both
systems' round-3 retrievals, so this is not a retrieval-coverage
failure. Instead it is a \emph{context-dilution} failure of NMC:
because both a topic chain and an event chain matched the
\textsf{forbidden city} entity, round 1's prompt was inflated with
14 pre-distilled chain items (vs.\ $0$ for MMG-only) that all
described post-entry events. On a counting question, this skews the
backbone's prior toward ``one prominent encounter'' before the
discriminating evidence even arrives. This pattern is consistent
with the MM-Lifelong Both cell's $-3.4$\,pp drop in
Table~\ref{tab:coverage}: when both chain types fire and the
question is open-ended on a low-baseline category (Counting,
Causal Reasoning), chain content can over-anchor the backbone rather
than reinforce the retrieved evidence. We discuss the corresponding
mitigation (relevance-pruning the merged context after each round)
under future work in Sec.~\ref{sec:limitations}.

\newpage
\section{Prompt templates}
\label{sec:prompts}

We list the system prompts for the load-bearing stages of the
MAGIC-Video pipeline, in the order they fire: captioning, MMG
construction, NMC distillation, and online inference. Each prompt
is followed by the
controller / user-role template at runtime; we omit the user-role
templates here because they are mechanical (they interpolate the
question, the retrieved context, or the running round history into a
single slot referenced in the system prompt). The MM-Lifelong Month
variants of the captioning and NMC prompts differ only in the narrator
preamble (IShowSpeed livestream rather than the A1\_JAKE household);
we include the EgoLifeQA variant here as the canonical form. The
\emph{final answerer} prompt, however, has two structurally different
variants because EgoLifeQA / Ego-R1 are multiple-choice while
MM-Lifelong is open-ended; both variants are listed verbatim below.

\subsection{Caption merge (ASR + visual, to a 30-s unified caption)}
Used in the MM-Lifelong captioning pipeline to fuse the Whisper transcript of a 30-second window with the per-frame VLM caption into a single coherent 30-s caption.

\begin{lstlisting}[style=prompt]
You are an expert video captioner. Your task is to merge a visual description and a speech transcript into a single, coherent caption.

# Input
You will receive:
- **Visual**: A description of what is visually happening in the video segment.
- **Transcript**: What is being said/spoken during the same segment.

# Guidelines
1. Combine both sources into one fluent paragraph.
2. Keep the visual description as the primary narrative.
3. Integrate speech content naturally (e.g., "The narrator explains '...' while the camera shows").
4. If transcript is empty, return the visual description as-is.
5. If visual description is empty but transcript exists, describe the speech.
6. Be concise: 2-4 sentences.

# Output
Output ONLY the merged caption text. No JSON, no explanations.
\end{lstlisting}

\subsection{Multi-granularity caption aggregation (30\,s, to 3\,min / 10\,min / 1\,h)}
Used to roll up the fine 30-s captions into coarser Episode captions at 3-min, 10-min, and 1-h granularities.

\begin{lstlisting}[style=prompt]
"You will be provided with some descriptions. 
Merge events into one single event based on these descriptions. 
Do not include uncertain information, speculation, or divergent content. 
Do not describe the atmosphere or emotions. Dismiss those provided content that are abstract or ambiguous. 
If the descriptions mention names of people who interacted with 'me', make sure to retain this information. 
Directly provide the summarized main events without adding any additional remarks or explanations.
\end{lstlisting}

\subsection{NER and semantic-triple extraction (combined OpenIE)}
A single LLM call that extracts named entities and schema-less $(s,p,o)$ triples from one episode caption. Produces the initial Entity and Semantic-triple nodes of the MMG.

\begin{lstlisting}[style=prompt]
Your task is to extract named entities and construct an RDF (Resource Description Framework) graph from the given paragraph.

Respond with a JSON object containing two keys:
1. "named_entities": A list of named entities found in the paragraph.
2. "triples": A list of RDF triples, each as a 3-element list [subject, predicate, object].

Pay attention to the following requirements:
- Each triple should contain at least one, but preferably two, of the named entities.
- When resolving pronouns, if the pronoun refers to the first-person (e.g., I, me, my), keep it as "I" instead of replacing with terms like "speaker" or "narrator". For other pronouns, clearly resolve them to their specific names to maintain clarity.
\end{lstlisting}

\subsection{Semantic-triple consolidation}
After triple extraction, this prompt decides for each new triple whether it is a duplicate, a contradiction, or a novel fact relative to existing triples at previous timestamps.

\begin{lstlisting}[style=prompt]
You are tasked with consolidating semantic knowledge by processing new semantic triple against relevant existing knowledge from previous timestamps.

Your job is to make two decisions:
1. **Which existing triples to remove/pop** - those that should be merged with the new triple or conflict with it
2. **How to update the new triple** - to reflect the consolidation, merge information, or resolve conflicts

# Consolidation Rules:
1. **Merge Similar Information**: If existing triples express very similar information to the new triple, remove them and update the new triple to capture the most complete/accurate representation
2. **Resolve Conflicts**: If the new triple conflicts with existing ones, decide which is more accurate/recent and remove the outdated ones
3. **Update with Context**: Use information from existing triples to make the new triple more specific or accurate
4. **Preserve Unique Information**: Only remove existing triples if they are redundant or conflicting

# Output Format:
Return ONLY a JSON object with the following two keys:
- `updated_triple` (List[str]): The new triple, possibly updated [subject, predicate, object]
- `triples_to_remove` (List[int]): Indices of existing triples to remove (empty list if none)
\end{lstlisting}

\subsection{NMC topic chain: per-entity biography extraction}
\label{sec:prompt_topic_chain}
Stage 3 of the topic-chain pipeline (candidate selection $\to$ caption collection $\to$ fact extraction). Applied once per surviving entity, producing a chronologically ordered list of timestamped, person-grounded facts. The MM-Lifelong variant replaces the A1\_JAKE narrator preamble with an IShowSpeed livestream description; all other instructions are identical.

\begin{lstlisting}[style=prompt]
In a first-person video, the entity "{entity}" appears in multiple episodes at different times.
The wearer is A1_JAKE (also referred to as "I"), living with housemates: Shure, Katrina, Tasha, Lucia, Alice, and sometimes visitors (Nicous, Choiszt, Jack, Violet, Lucy).

Below are the 30-second episode captions where "{entity}" is mentioned:

{episodes_text}

Extract facts about "{entity}" that help answer these types of questions:

1. **Who & When** (EntityLog): Who used/touched/moved "{entity}" first? Whose "{entity}" is this? Where was "{entity}" before?
2. **What happened** (EventRecall): When was "{entity}" first/last mentioned or discussed? What happened with "{entity}" last time?
3. **Relationships** (RelationMap): Who helped whom with "{entity}"? Who gave/passed "{entity}" to whom? Who was together when using "{entity}"?
4. **Plans & Decisions** (TaskMaster): Who suggested/planned to buy/make/use "{entity}"? What was decided about "{entity}"?
5. **Habits & Preferences** (HabitInsight): Who always/usually/often uses "{entity}"? Who likes/dislikes "{entity}"? What does someone typically do with "{entity}"?

Rules:
- Always use specific person names (A1_JAKE, Shure, Katrina, Tasha, Lucia, Alice, Nicous, Choiszt, Jack). NEVER use "I", "he", "she", "they", "we".
- Include the WHERE (which room, location) when relevant.
- Record preferences and habits ("Tasha refused the {entity}", "Shure always drinks {entity} in the morning").
- Record plans and discussions ("They decided to order {entity} for lunch", "A1_JAKE suggested buying {entity}").
- Skip trivial mentions where "{entity}" appears in background without meaningful action.

Good facts:
- "Shure handed the {entity} to A1_JAKE in the kitchen"
- "Lucia and A1_JAKE set up the {entity} together in the living room"
- "Katrina said she doesn't like {entity}"
- "A1_JAKE suggested ordering {entity} for dinner, everyone agreed"
- "Tasha was the first person to use the {entity} this morning"

Bad facts (skip these):
- "A1_JAKE picked up the {entity}" (no meaningful interaction)
- "The {entity} was on the table" (static description, no action)

Output a JSON array (empty array [] if no meaningful facts):
[
  {{"time": "DAY1 17:45", "fact": "descriptive fact with person names, location, and action"}},
  {{"time": "DAY2 11:00", "fact": "another fact"}}
]
\end{lstlisting}

\subsection{NMC event chain --- Step 1 (per-day activities) and Step 2 (cross-time chain discovery)}
\label{sec:prompt_event_chain_12}
Step 1 feeds the 10-minute captions of a single day to an LLM and asks for a list of major activities. Step 2 concatenates the per-day activity lists across the entire recording and asks a second LLM pass to identify related activities across time (recurring routines, progressive project stages, intermittent work). A second round of Step 2 asks the LLM to find chains missed by the first round.

\begin{lstlisting}[style=prompt]
=== STEP1_PROMPT ===
You are analyzing a first-person daily life video. The narrator is A1_JAKE, living with housemates: Shure, Katrina, Tasha, Lucia, Alice, and sometimes visitors (Nicous, Choiszt, Violet, Jack, Liu Ting).

Below are 10-minute summary segments from {date}.

For each major activity that happened on this day, extract:
- name: a short descriptive name (e.g., "Making a cake", "Shopping at supermarket")
- time_range: approximate start and end time (e.g., "11:00:00-12:30:00")
- summary: 2-3 sentences describing what happened. Must include:
  * WHO specifically participated (by name, never "he/she/they/I/we")
  * WHO helped whom, WHO worked together
  * WHAT concrete actions were taken (not vague words like "coordinated", "handled", "managed")
  * WHAT was the outcome or decision made
- key_entities: important objects or places involved (do NOT include person names here --- only objects and places)

Return a JSON list:
[
  {{
    "name": "activity name",
    "time_range": "HH:MM:SS-HH:MM:SS",
    "summary": "detailed description with person names, specific actions, and outcomes",
    "key_entities": ["object1", "place1"]
  }}
]

BAD summary: "A1_JAKE coordinated box stacking and workspace assembly planning with the group."
GOOD summary: "A1_JAKE and Shure carried boxes upstairs and assembled a display rack. Lucia asked about placement and helped hold the frame. Katrina suggested making the rack longer. The rack was installed with a whiteboard on top."

Focus on significant, goal-directed activities (cooking, shopping, meetings, performances, crafts), not trivial actions. Aim for 5-15 activities per day.

Segments:
{segments}

=== STEP2_PROMPT ===
Below are daily activities from a 7-day first-person video. The participants are: A1_JAKE, Shure, Katrina, Tasha, Lucia, Alice, and visitors Nicous, Choiszt.

Find activities that are related across different time periods. This includes:
- Different stages of the same project (e.g., "discussed cake plan" DAY1 -> "bought ingredients" DAY3 -> "baked the cake" DAY5)
- Repeated occurrences of the same activity (e.g., "brewed coffee" on DAY2, DAY3, DAY4)
- Ongoing work done intermittently (e.g., "room decoration" on DAY1 morning and DAY2 evening)
- Same type of errand repeated (e.g., "shopping at supermarket" on DAY1, DAY3, DAY5)

Only output events that have 2 or more related activities across different time periods. Ignore standalone activities. Be thorough and find as many as possible.

For each event, output:
- name: specific name
- steps: list of {{day, start_time, end_time, description}}
  * description must name specific people involved (never "he/she/they/I/we")
  * description must state concrete actions and outcomes, not vague summaries
  * description should explain how this step relates to the previous step (continuation, escalation, completion, etc.)
- key_entities: important objects and places only (do NOT include person names)

BAD description: "A1_JAKE handled coffee preparation while the group reviewed the shopping list."
GOOD description: "Shure brewed pour-over coffee for everyone in the courtyard. A1_JAKE served espresso shots. Lucia and Katrina discussed which coffee beans to buy for the next day."

Return JSON list.

Daily activities:
{daily_activities}
\end{lstlisting}

\subsection{NMC event chain --- Step 3 (step enrichment)}
\label{sec:prompt_event_chain_3}
For each step of each discovered chain, this prompt reads the 30-s captions falling inside the step's time range and asks the LLM to produce a detailed, event-specific step description, replacing the coarser Step-2 summary.

\begin{lstlisting}[style=prompt]
You are analyzing a first-person daily life video. The narrator is A1_JAKE, living with housemates: Shure, Katrina, Tasha, Lucia, Alice, and sometimes visitors (Nicous, Choiszt).

This is one step of the event chain "{chain_name}".
Time: {day} {start_time} - {end_time}

Below are the 30-second episode captions from this time period. Extract ONLY information relevant to "{chain_name}". Ignore unrelated content.

Captions:
{captions_text}

Write 2-5 sentences describing what happened in this step, focusing on:
- WHO specifically participated (by name, never "I/he/she/they/we")
- WHO helped whom, WHO gave what to whom
- WHAT concrete actions were taken and what was the outcome
- WHAT decisions were made

Output a single JSON string (just the description text):
{{"description": "detailed description here"}}
\end{lstlisting}

\subsection{Online retrieval controller (MAGIC-Video, agentic \texttt{search}/\texttt{answer} loop)}
\label{sec:prompt_controller_mmg}
The LLM-controller prompt for MAGIC-Video's agentic loop (\emph{+ MMG} and \emph{+ NMC} configurations of Figure~\ref{fig:ablation}). At each round, the controller receives the question and the current round history, and emits a JSON decision that is either \texttt{search} (with a natural-language \texttt{search\_query}) or \texttt{answer}. The user-role template (omitted here) simply formats the running round history into the slot referenced in the system prompt.

\begin{lstlisting}[style=prompt]
You are a reasoning agent for a multimodal video memory retrieval system.
Your job is to decide whether to stop and answer, or to search memory for more evidence.

# Decision Modes:
1. **search**: Retrieve memory to begin, continue, or extend progress toward the answer.
   - Write the search query as a **natural-language sentence or question** (NOT a list of keywords).
     Good:  "Who handed the black marker to Shure?"
     Bad:   "black marker Shure hand location"
   - The retrieval system uses semantic embedding similarity, so natural sentences work much better than keyword lists.
   - Each round, try a **different angle** --- do not rephrase the same query.
2. **answer**: Stop searching because the accumulated results are sufficient.
   - If 2+ consecutive rounds returned "[No new results]", you MUST answer with what you have.

# Context Inputs:
- Current Query
- Round History: Log of past retrieval rounds. Each round is written in this format:

  ### Round N
  Decision: <search|answer>
  Search Query: <query text>
  Retrieved:
  <retrieved items summary>

# STRICT OUTPUT RULES:
- Always decide **first**: "search" or "answer".
- If decision = "search": Must include "search_query" (a single concise query string).
- If decision = "answer": Do NOT include "search_query".
- Always output valid JSON only, no extra commentary.

# Output Format:
{
  "decision": "search" | "answer",
  "search_query": "<str>"
}

# Few-shot Examples:
## Example 1
Query: Who gives the graduation gift to Maria?
Round History: []

### Response:
{
  "decision": "search",
  "search_query": "Who gave a graduation gift to Maria?"
}

## Example 2
Query: Who gives the graduation gift to Maria?
Round History:
### Round 1
Decision: search
Search Query: Who gave a graduation gift to Maria?
Retrieved:
[DAY1 10:30:00 - DAY1 10:31:30] (30sec)
I watch Luis hand a wrapped gift to Maria at the ceremony.
(Luis, gives, graduation gift to Maria)

### Response:
{
  "decision": "search",
  "search_query": "What is Luis's relationship to Maria?"
}

## Example 3
Query: Who gives the graduation gift to Maria?
Round History:
### Round 1
Decision: search
Search Query: Who gave a graduation gift to Maria?
Retrieved:
[DAY1 10:30:00 - DAY1 10:31:30] (30sec)
I watch Luis hand a wrapped gift to Maria at the ceremony.

### Round 2
Decision: search
Search Query: What is Luis's relationship to Maria?
Retrieved:
(Luis, is brother of, Maria)

### Response:
{
  "decision": "answer"
}

## Example 4 (incorporate discovered clues into later queries)
Query: Who used the microwave last on the first floor?
Round History:
### Round 1
Decision: search
Search Query: Who used the microwave on the first floor?
Retrieved:
[DAY1 20:32:00 - DAY1 20:32:30] (30sec)
Lucia asks, "Can it fit?" I reply, "Yes." I put a plate in the microwave.
(Lucia, adjusts, microwave)
(I, uses, microwave)

### Response:
{
  "decision": "search",
  "search_query": "Did Lucia or anyone else use the first-floor microwave after DAY1 20:32?"
}

## Example 5 (no new results --- stop early)
Query: Where did I put the red box?
Round History:
### Round 1
Decision: search
Search Query: Where did I place the red box?
Retrieved:
[No new results]

### Round 2
Decision: search
Search Query: What happened with the red box recently?
Retrieved:
[No new results]

### Response:
{
  "decision": "answer"
}
\end{lstlisting}

\subsection{Online retrieval controller (Independent Retrieval baseline, WorldMM-style)}
\label{sec:prompt_controller_ir}
The IR baseline of Figure~\ref{fig:ablation} uses WorldMM's~\citep{worldmm} original controller prompt verbatim, since the IR row is a faithful re-implementation of WorldMM's pipeline. It is structurally different from the MAGIC-Video prompt above in two ways: (i) at each search round it requires the agent to \emph{select one memory type} (\texttt{episodic} / \texttt{semantic} / \texttt{visual}) and route the query to that single index, and (ii) it instructs the agent to emit a \emph{keyword-style} search query (rather than a natural-language sentence), which matches WorldMM's underlying BM25-driven per-modality top-$k$ retrieval. We do not use this prompt for our \emph{+ MMG} / \emph{+ NMC} configurations because cross-modal PPR returns a mixed node-type set in a single pass (no per-modality routing needed) and dense-embedding seeding plus NMC's Tier-2 cosine matching both work substantially better with sentence-form queries.

\begin{lstlisting}[style=prompt]
You are a reasoning agent for a video memory retrieval system.
Your job is to decide whether to stop and answer, or to search memory for more evidence.
When searching, you must select exactly one memory type and form a query.

# Decision Modes:
1. **search**: Retrieve memory to begin, continue, or extend progress toward the answer
   - Choose one memory type and form a keyword(phrase)-style search query.
2. **answer**: Stop searching because the accumulated results are sufficient.
   - No memory type selection is needed.

# Memory Types:
1. Episodic: Specific events/actions. Query by EVENT/ACTION.
2. Semantic: Entities/relationships. Query by ENTITY/CONCEPT.
3. Visual: Scene/setting snapshots. Query by SCENE/SETTING or TIMESTAMP RANGE.

# STRICT OUTPUT RULES:
- Always decide **first**: "search" or "answer".
- If decision = "search": Must include "selected_memory" with exactly one memory type and one query.
- If decision = "answer": Do NOT include "selected_memory".
- Always output in valid JSON only, no extra commentary.

# Output Format:
{
  "decision": "search" | "answer",
  "selected_memory": {
    "memory_type": "episodic" | "semantic" | "visual",
    "search_query": <str>
  } # Omit if decision = "answer"
}

# Few-shot Examples (abridged):
## Example 1
Query: Who gives the graduation gift to Maria?
Round History: []
### Response:
{ "decision": "search",
  "selected_memory": {"memory_type": "episodic",
                      "search_query": "Maria graduation gift giver"} }
\end{lstlisting}

\subsection{Final answerer (chain-augmented QA, EgoLifeQA / Ego-R1, multiple choice)}
Once the agentic loop commits to \texttt{answer}, the accumulated multi-round context (episode captions, semantic triples, visual frames, and NMC-injected narrative facts) is passed to this prompt, which produces the final multiple-choice answer for EgoLifeQA and Ego-R1.

\begin{lstlisting}[style=prompt]
You are an AI assistant that answers questions about egocentric video experiences using retrieved memory context. Your task is to answer multiple choice questions based on this accumulated context. Always choose the most relevant answer from the given choices based on the evidence provided.

# Context Types
Your context contains different types of information:
- [Retrieved episode]: Video segments retrieved by relevance search.
- [Topic: X]: Key facts about entity X tracked across the entire video timeline.
- [Event: X]: Steps of a recurring activity X across multiple days.
- [Retrieved semantic]: Knowledge triples extracted from the video.
- [Visual frames]: Video frames from retrieved episodes.

# Guidelines
- Analyze all provided context carefully.
- Pay attention to [Topic] and [Event] facts --- they contain information from other time periods that the retrieved episodes may not cover.
- Choose the answer that best matches the evidence.
- If evidence is unclear, make the most reasonable inference.

# Output Format
Provide your answer as a single letter (A, B, C, or D) based on the evidence.
\end{lstlisting}

\subsection{Final answerer (chain-augmented QA, MM-Lifelong, open-ended)}
\label{sec:prompt_answerer_mml}
The MM-Lifelong variant of the answerer drops the multiple-choice constraint and instructs the model to produce a free-form direct response. Selected at runtime via \texttt{qa\_template\_name = "qa\_mmlifelong"} (chain mode: \texttt{qa\_mmlifelong\_chain}).

\begin{lstlisting}[style=prompt]
You are an AI assistant that answers questions about livestream video broadcasts using retrieved memory context. Your task is to answer open-ended questions based on the accumulated context.

# Context Types
Your context contains different types of information:
- [Retrieved episode]: Video segments retrieved by relevance search.
- [Topic: X]: Key facts about entity X tracked across the video timeline.
- [Event: X]: Steps of a recurring activity X across the broadcast.
- [Retrieved semantic]: Knowledge triples extracted from the video.
- [Visual frames]: Video frames from retrieved episodes.

# Guidelines
- Analyze all provided context carefully.
- Pay attention to [Topic] and [Event] facts --- they contain information from other time periods that the retrieved episodes may not cover.
- If the question asks for a count, give the number.
- If the question asks for an ordering, give the correct sequence.
- If the question asks to identify items from a list, state which ones are correct.
- If evidence is unclear, make the most reasonable inference.
- Keep answers brief and factual (1-2 sentences unless the question requires more detail).

# Output Format
Provide your answer as a direct response to the question. Do NOT format as multiple choice.
\end{lstlisting}

%%%%%%%%%%%%%%%%%%%%%%%%%%%%%%%%%%%%%%%%%%%%%%%%%%%%%%%%%%%%

\end{document}